\pdfoutput=1
\documentclass{article}

\usepackage[preprint]{neurips_2026}

\usepackage[utf8]{inputenc}
\usepackage[T1]{fontenc}
\usepackage{hyperref}
\usepackage{url}
\usepackage{booktabs}
\usepackage{amsfonts}
\usepackage{amsmath}
\usepackage{nicefrac}
\usepackage{microtype}
\usepackage{xcolor}
\usepackage{graphicx}
\usepackage{longtable}
\usepackage{multirow}

\hypersetup{pdftitle={Language Unalignability: Why Some Concepts Resist Cross-Cultural Benchmark Evaluation},pdfauthor={Shu-Kai Hsieh, Da-Chen Lian}}

\title{Language Unalignability: Why Some Concepts Resist Cross-Cultural Benchmark Evaluation}

\author{%
  Shu-Kai Hsieh\\
  National Taiwan University\\
  \texttt{shukaihsieh@ntu.edu.tw}
  \And
  Da-Chen Lian\\
  National Taiwan University\\
  \texttt{d08944019@ntu.edu.tw}
}

\begin{document}

\maketitle

\begin{abstract}

Current evaluation of multilingual Large Language Models (LLMs) rests on an implicit Translation-Isomorphism Assumption (TIA): that semantic structures across languages are fundamentally congruent and mutually mappable without loss of information. We argue that this assumption is not merely violated in practice, but ill-posed in principle for a typologically identifiable class of concepts, including pragmatic markers, honorifics, and diachronically stratified terms. We formalize this failure using a usage-cloud framework, representing concepts as point sets of contextualized embeddings. We define $\alpha$-unalignability as the mathematical impossibility of any mapping that simultaneously preserves lexical faithfulness (centroid correspondence) and structural faithfulness (local neighborhood topology). We provide three layers of evidence for this phenomenon. Behaviorally, we show that FLORES-200 translation failures are predicted by language family and resource class but not by script, a proxy for surface exposure, and that LOBSTER reasoning scores vary by language family. Mechanistically, we report a Representation–Intervention Gap (RIG) in a nine-model case study on Yami: the models' activations encode a regularity along which Yami groups with other low-resource and Austronesian languages, yet interventions on language-specific neurons show no demonstrated advantage over random masks—the regularity is visible but not usable by this intervention. Finally, we operationalize these findings into a multidimensional diagnostic profile, incorporating Cycle-Consistency, Pragmatic-Load Disagreement, Manifold-Curvature Mismatch, and RIG. We argue that collapsing cultural competence into a single scalar metric incentivizes ``probabilistic flattening,'' and that recognizing the unalignable class is a necessary precondition for AI that respects, rather than erases, cultural divergence. This suggests that multilingual alignment is not a single well-defined objective, but a set of mutually incompatible projections.

\end{abstract}

\section{Introduction}\label{sec:intro}

Current multilingual large language models are more fluent than any
prior system and more systematically wrong about the social-pragmatic
content of what they translate. A model can render \emph{x\={\i}n
k\v{u} le} (Mandarin: ``you've worked hard'') as \emph{good job} in
idiomatic English across all of its attested uses
(superior-to-subordinate, peer-to-peer, addressing a service worker,
ironic) and receive identical scores on every leading multilingual
benchmark, since such benchmarks score against one or a few fixed
references per source item. The benchmark architecture itself enforces a collapse of
pragmatically distinct acts: the failure lies in the evaluation
paradigm. This flattening is not a transient defect but the symptom of
a structural assumption cross-cultural NLP has not articulated and
therefore cannot test.

We name this assumption the \textbf{Translation-Isomorphism Assumption}
(TIA): that for any concept $C$ attested in language $L_A$, there exists a translation
equivalent $C'$ in $L_B$ such that the meaning-preserving map
$C \mapsto C'$ is, modulo idiomatic noise, a function whose inverse is also meaning-preserving. TIA is the precondition under which gold standards are constructible.
Cross-lingual word-embedding research has chipped at it at the lexical
level for nearly a decade \citep{sogaard-etal-2018-limitations,vulic-etal-2020-good}; we generalize the failure from word vectors to the contextualized \emph{clouds} generated by modern encoders, and argue that for a typologically identifiable class of concepts (pragmatic markers, honorifics, evidentials, diachronically loaded terms) no
faithful map exists, in a sense we make precise.

Our thesis: \textbf{unalignability} is a structural geometric property
of certain concepts, not an anthropological complaint about
translation, with three observable consequences. \textbf{First},
documented behavioral failures in cross-cultural-NLP benchmarks
systematically violate a structural-faithfulness condition that
centroid-style scoring cannot see. \textbf{Second}, those failures
pattern along typological as well as resource lines: in separate one-way
analyses on FLORES-200, language family is a predictor of translation
quality comparable to or stronger than resource class, while
\emph{script is not} a significant predictor
\citep{lian-etal-2025-lobster}. \textbf{Third}, we hypothesize that unalignability has a
mechanistic signature inside the model: a \emph{Representation--Intervention
Gap} (RIG), in which a regularity recoverable from activations is not
usable by localization-based intervention.

The paper develops these claims in sequence. \S\ref{sec:frame} introduces the
cloud-mapping framework. \S\ref{sec:behav} re-interprets findings from FLORES-200 and LOBSTER as systematic structure-faithfulness violations. \S\ref{sec:rig} presents the mechanistic argument via RIG. \S\ref{sec:metrics} operationalizes the framework into a four-metric diagnostic profile. \S\ref{sec:rec}--\S\ref{sec:disc} discuss consequences.

We are not arguing that translation between languages is globally
indeterminate, nor that low-resource performance is structurally
unimprovable. The position is more specific: that \emph{some concepts
are structurally unalignable}, that the class admits typological
characterization, and that current evaluation paradigms penalize models
for failures that are, in part, properties of the evaluation setup
itself.

\section{Background}\label{sec:bg}

\paragraph{Cross-lingual representation alignment.}
Cross-lingual word-embedding research has formally tested the
translation-isomorphism intuition for nearly a decade.
\citet{mikolov2013exploitingsimilaritieslanguagesmachine} proposed
bilingual lexicon induction as a linear map between monolingual spaces,
motivating VecMap \citep{artetxe-etal-2018-robust}, MUSE
\citep{lample2018word}, and a substantial subsequent literature.
Follow-up work systematically dismantled the assumption:
\citet{sogaard-etal-2018-limitations} showed isomorphism fails between
genealogically close languages; \citet{vulic-etal-2020-good}
extended the negative result across a typologically diverse sample,
while attributing much of it to monolingual resource size and training
regime rather than typology alone;
\citet{patra-etal-2019-bilingual} demonstrated weak-isomorphism only
under restrictive distributional assumptions. Our position generalizes
this thread: type-level failure is a special case of a deeper failure
at the level of \emph{contextual usage clouds}, where pragmatic load
and diachronic stratification compound the geometric mismatch.

\paragraph{Cross-cultural NLP evaluation.}
The first generation of cross-cultural benchmarks (e.g., TMLU \citep{chen2024measuring}, TMMLU+ \citep{ikala2024improved},
CulturalBench \citep{chiu-etal-2025-culturalbench}, and TaiwanVQA \citep{hsieh-etal-2025-taiwanvqa}) applied trivia-style scoring to
culturally localized content. Recent critiques converge on a Geertzian
distinction: these instruments measure \emph{thin} surface knowledge
rather than \emph{thick} situated competence \citep{vo2025cureculturalunderstandingreasoning}.
A second wave of \emph{bottom-up emic} benchmarks (NORMSAGE
\citep{fung-etal-2023-normsage}, BLEnD \citep{myung2024blend}, CARE
\citep{guo2025care}, and the BETEL pilot illustrated in
\S\ref{sec:behav}) inductively extracts probe items from native-speaker
practice, targeting situated pragmatic competence rather than
encyclopedic recall. Our framework engages this second family directly:
the discriminating quantity in their probes (pragmatic appropriateness
conditioned on situational and historical context) is operationally
the structure-faithfulness condition (M2) of \S\ref{sec:frame}.

\paragraph{Mechanistic interpretability and causal accessibility.}
The mechanistic argument of \S\ref{sec:rig} sits in the lineage of
representation probing \citep{belinkov2022probing}, causal abstraction and Distributed
Alignment Search \citep{geiger2024finding}, sparse autoencoder feature
decomposition \citep{bricken2023towards,huben2024sparse}, and
language-specific neuron identification via LAPE
\citep{tang2024languagespecificneurons}. A recurring methodological
tension: what activations \emph{encode} (probe-decodable) need not
coincide with what they \emph{cause} (intervention-effective). Our
Representation--Intervention Gap formalizes this dissociation, well
known at the syntax and entity-coreference levels. Our specific
contribution is to argue that the dissociation has its sharpest
empirical instance at the level of \emph{concept-class unalignability
across languages and time}, where the structure to be intervened on is
precisely the structure that resists single-locus localization.

\section{The geometric frame}\label{sec:frame}

\subsection{The translation-isomorphism assumption}

Almost every cross-lingual evaluation paradigm rests on an assumption so
basic it is rarely stated. We name it the
\textbf{Translation-Isomorphism Assumption} (TIA):

\begin{quote}
For any concept $C$ attested in language $L_A$, there exists a translation
equivalent $C'$ in $L_B$ such that the meaning-preserving
mapping $C \mapsto C'$ is, modulo idiomatic noise, a function whose inverse exists and is
meaning-preserving as well.
\end{quote}

TIA is the precondition under which \emph{gold standards} are
constructible. If $C \mapsto C'$ is not a function (because it depends
ineliminably on context, register, speech-act conditions, or
\emph{historical period}), no single reference translation is
\emph{the} correct one, and BLEU-style evaluation against a fixed gold
misrepresents the task. The literature reviewed in \S\ref{sec:bg} documents TIA's failure
at the lexical level; we claim it is a special case of a broader
failure at the level of \emph{contextual usage}, which current
benchmark designs continue to assume implicitly even after the lexical
critiques. TIA is rarely stated explicitly, yet it is the hidden axiom
under which all cross-lingual evaluation is constructed.

\subsection{Concepts as usage clouds}

The unit of comparison should not be the lexical type. Pragmatic
markers, honorifics, evidentials, and diachronically loaded terms (the loci of cross-cultural mistranslation) derive their meaning from
the \emph{distribution} of contexts in which they are deployed, not from
a fixed denotation. We propose to identify a concept with the \emph{cloud} of contextualized embeddings it generates.

\textbf{Definition 1 (Usage Cloud).} \emph{Let $\mathcal{M}$ be a
multilingual encoder mapping (text, position) pairs to vectors in $\mathbb{R}^d$. For a
target token sequence $C$ attested in language $L$ over a time window $t$, let $\mathcal{X}_L(C, t) = \{x_1, \ldots, x_K\}$ be a corpus of attested sentences containing $C$ in that window. The \textbf{usage cloud} of $C$ is}
\[
\Phi_L(C, t) \;=\; \big\{\, \mathcal{M}(x_i, \mathrm{pos}(C \text{ in } x_i)) \;\big|\; x_i \in \mathcal{X}_L(C, t)\,\big\} \;\subset\; \mathbb{R}^d.
\]

Note that the cloud is a \textbf{point set}, not a manifold. We make no Riemannian commitments here; what is well-defined on a point set suffices: pairwise distances, intrinsic dimensionality, density profiles, anisotropy, and residuals of attempted mappings. The intrinsic dimensionality $d_{int}$ of $\Phi_{L}(C,t)$ serves as a proxy for its pragmatic complexity; we hypothesize that ``unalignable'' concepts exhibit dimensionality expansion in languages where these categories are grammatically mandatory. Three properties matter.

\emph{Pragmatic load is geometric load.} The four registers of
\emph{x\={\i}n k\v{u} le} (\S\ref{sec:intro}) occupy distinguishable
regions of $\Phi_{\mathrm{cmn}}(\text{x\={\i}n k\v{u} le})$, separated
by directions encoding social register and footing in
\citeauthor{goffman1979footing}'s sense. English \emph{worked hard}
occupies a comparatively low-dimensional region; Mandarin \emph{x\={\i}n k\v{u} le}
expands into additional dimensions encoding social hierarchy and
debt-reciprocity. Pragmatic load adds orthogonal directions to the
usage cloud that cannot be collapsed without losing structural
faithfulness (M2).

\emph{Diachrony is non-stationarity.} The cloud is necessarily
time-indexed: the same lexical type in the same language can occupy
systematically different regions of the embedding space at different
historical periods (\S\ref{sec:diachrony}).

\emph{Cross-lingual comparison is cross-cloud comparison.} The translation pair $(C, C')$ corresponds to two clouds, $\Phi_A(C, t)$ and $\Phi_B(C', t)$. Whether the translation is \emph{adequate} reduces to whether the two clouds stand in some structure-preserving relation, a question we now make precise.

\subsection{Alignment as cloud mapping}

\textbf{Definition 2 (Alignment Map).} \emph{Given clouds $\Phi_A(C, t)$ and $\Phi_B(C', t)$, an alignment map $\tau : \Phi_A \to \Phi_B$ is \textbf{lexically faithful} if it preserves cloud centroids,}
\begin{equation}\label{eq:m1}
\big\|\,\tau(\mu_A) - \mu_B\,\big\|_2 \;<\; \varepsilon, \qquad \mu_L = \tfrac{1}{|\Phi_L|}\sum_{v \in \Phi_L} v, \quad \textup{(M1)}
\end{equation}
\emph{and \textbf{structurally faithful} if it preserves local
neighborhood ranks,}
\begin{equation}\label{eq:m2}
\frac{1}{|\Phi_A|}\sum_{v \in \Phi_A}\big| \mathrm{rank}_k(v; \Phi_A) - \mathrm{rank}_k(\tau(v); \tau(\Phi_A)) \big| \;<\; \delta. \quad \textup{(M2)}
\end{equation}

(M1) is tacitly enforced by every dictionary-style benchmark: \emph{centroids} coincide under translation. All commercially deployed translation systems satisfy it for high-frequency vocabulary. (M2) is the more demanding condition: that the \emph{internal contrastive structure} of $\Phi_A(C)$ (the pragmatic, register, and diachronic differentiations distinguishing its sub-regions) survives under $\tau$. Our position: existing benchmarks fail to test (M2), and a class of culturally loaded concepts systematically resists it. We choose rank-based over distance-based faithfulness deliberately: distance-based criteria conflate genuine structural change with the global anisotropy of contextual-embedding spaces \citep{ethayarajh-2019-contextual}. Rank-based criteria measure what we actually care about: whether the \emph{ordering} of contextual contrasts survives under $\tau$.

Crucially, geometric faithfulness at the level of $\tau$ does not guarantee \emph{operational usability} under intervention, a distinction we operationalize in \S\ref{sec:rig}.

The formalism contacts three traditions. Procrustes-style cross-lingual mapping \citep{lample2018word} optimizes (M1) under orthogonality constraints. Optimal-transport approaches \citep{zhang-etal-2017-earth} interpolate between (M1) and (M2) by penalizing point-mass redistribution. Distributed Alignment Search \citep{geiger2024finding} reads, in our terms, as the search for a $\tau$ satisfying (M2) at the \emph{causal} rather than the \emph{correlational} level, leveraged in \S\ref{sec:rig}.

\subsection{Diachrony as a stress test}\label{sec:diachrony}

\emph{Within a single language, a concept can fail to align with itself
across time.} Take the Mandarin term \textit{q\'{i}ng}, glossed today as `feeling, emotion, affection' but in classical usage centered on `(factual) circumstance, situation, true state of affairs'. Compare $\Phi_{\mathrm{cmn}}(\textit{q\'{i}ng}, t_1)$ for the Pre-Qin to Han period with $\Phi_{\mathrm{cmn}}(\textit{q\'{i}ng}, t_2)$ for Modern Mandarin: same lexical type, same
language, same encoder; yet centroids diverge and sub-region structure reorganizes. Any $\tau$ attempting to align the two must violate either (M1) or (M2). TIA thus presupposes not just a single language but a single \emph{time slice} of that language, an even stronger and even less-defended assumption.

Models carry an implicit synchronic prior: training is overwhelmingly post-2000 web text, so $\Phi(C, t_{\mathrm{train}})$ is a biased estimator of $\Phi(C, t)$ for any earlier $t$. When a benchmark uses contemporary glosses to score model output on classical, religious, or historical text, the gold standard encodes a synchronic flattening the evaluation cannot detect. In terms of the metrics of \S\ref{sec:metrics}, the diachronic case would be the cleanest empirical test of Cycle-Consistency (CC) and Manifold-Curvature Mismatch (MCM): same encoder, same language, same lexical type, two time windows, no language-distance, resource-asymmetry, or tokenization confounds.

\subsection{Unalignability as structural distortion}

\textbf{Definition 3 ($\alpha$-Unalignability).} \emph{A concept pair $(C, C')$ is $\alpha$-unalignable, given encoder $\mathcal{M}$ and tolerance vector $\alpha = (\varepsilon, \delta)$, iff}
\begin{equation}\label{eq:unalignable}
\inf_{\tau \,\in\, \mathrm{Maps}(\Phi_A, \Phi_B)}\; \max\!\Big\{\frac{\mathrm{viol}_{\textup{M1}}(\tau)}{\varepsilon},\; \frac{\mathrm{viol}_{\textup{M2}}(\tau)}{\delta}\Big\} \;>\; 1,
\end{equation}

i.e., no map satisfies both (M1) within $\varepsilon$ and (M2) within $\delta$. Three structurally different routes lead to a large infimum. They are not mutually exclusive (most culturally loaded concepts realize more than one), but are diagnostically separable.

\emph{Route~A: Pragmatic-load asymmetry.} The source cloud $\Phi_A(C)$ is partitioned by a discrete pragmatic dimension along which the target language $L_B$ lacks a corresponding dimension; any $\tau$ collapses the partition. Honorifics across honorific-poor targets, evidentials \citep{aikhenvald2004evidentiality} across non-evidential targets, and case-marked information structure across English-style fixed-word-order targets all sit here.

\emph{Route~B: Diachronic non-stationarity.} $\Phi_A(C, t)$ and $\Phi_A(C, t')$, same language but different time, differ enough that any single $C'$ in $L_B$ can match at most one. \emph{q\'{i}ng} across roughly two millennia is the canonical case.

\emph{Route~C: Categorial non-correspondence.} The clouds occupy categorially distinct regions of the model's representational space. The target cloud is not a deformation of the source cloud but a \emph{projection} onto a lower-dimensional subspace.

These three routes correspond to \emph{indexicality}, \emph{semantic change}, and \emph{lexical-field reorganization}, converging on a single ML observation: the loss surface of the search for $\tau$ is structurally barred from a low-distortion solution. Figure~\ref{fig:clouds} (Appendix~\ref{app:figure}) visualizes the framework, with the source--target asymmetry on top and the three routes below.

\paragraph{Philosophical positioning.}
The framework sits between Quine's indeterminacy thesis \citep{quine2013word} and Wierzbicka's Natural Semantic Metalanguage \citep{wierzbicka1996semantics}. We do not claim translation is \emph{globally} indeterminate; we claim the indeterminacy is \emph{concentrated} on a structurally identifiable class.

\section{Behavioral evidence}\label{sec:behav}

If concepts are usage clouds and alignment is cloud mapping, then a
class of concepts admits no low-distortion $\tau$, and current benchmarks
tacitly score on (M1) only. This section shows that documented failures are
systematic (M2) violations.

\subsection{The (M1)-only trap}

FLORES-200 \citep{nllbteam2022languageleftbehindscaling}, MEGA/MEGAVERSE \citep{ahuja2023mega,ahuja-etal-2024-megaverse}, XTREME-UP \citep{ruder-etal-2023-xtreme}, and Belebele \citep{bandarkar-etal-2024-belebele} define the empirical floor of contemporary multilingual evaluation. Each scores against one or a few reference translations under metrics (chrF, BLEU, COMET, multiple-choice accuracy) sensitive to \emph{centroid-level} lexical correspondence. None distinguishes adequacy on the dominant register of a target concept from adequacy on its sub-register structure: a model translating \emph{x\={\i}n k\v{u} le} uniformly as \emph{good job} across all four pragmatic contexts receives identical scores against a gold standard with the same flattening.

This is deliberate design, driven by the cost of authoring multiple registered translations. Our framework diagnoses it: the metrics operate on $\tau(\mu_A) \approx \mu_B$, silently substituting this for $\tau$ being a faithful map. \citet{hershcovich-etal-2022-challenges} characterize the resulting blind spots as a ``challenge''; we sharpen this to a structural prediction. Any benchmark whose reference is a single point per source item cannot, even in principle, detect (M2) violation.

\subsection{Script does not predict failure; family and resource do}

Three findings anchor the typological claim.

\textbf{Resource and family, but not script, predict translation
quality.} A recent preliminary FLORES-200 evaluation of Gemini-2.5-Flash (ten sentences per language, single run) across 204
languages \citep{lian-etal-2025-lobster} finds: (1) generation failure rate is roughly $10\times$ higher in the English-to-target direction than the reverse, suggesting the binding constraint is not in \emph{receiving} the source-side cloud but in \emph{projecting} into the target-side one; (2)~one-way ANOVAs find both language family
($\eta^2_p = 0.41$ for E$\to$T) and resource class ($\eta^2_p = 0.41$ for E$\to$T) highly significant predictors of chrF in both directions ($p < .001$); (3)~most diagnostic for our framework, \emph{script is not a significant predictor} ($\eta^2_p = 0.17$, $p = .27$ for E$\to$T). If failure were primarily orthographic (driven by tokenizer exposure to the surface form), script should matter; in this evaluation it does not, although the LOBSTER authors note that script is itself confounded with resource availability.

\textbf{Typological reasoning scores vary by family.} The same line of work introduces LOBSTER, a benchmark of 96 linguistic-reasoning problems from the International Linguistics Olympiad (IOL) across more than 90 languages with rich typological metadata. IOL puzzles are designed to be \emph{self-contained}, solvable from puzzle-internal data alone, although the LOBSTER authors note that models may still have some prior exposure to the languages. Gemini-2.5-Pro scores highest on language isolates (mean 0.70), Turkic (0.64), and Indo-European (0.55), and struggles with Papuan (0.29), Australian (0.34), and South American (0.25) languages. The LOBSTER authors note that this trend may be partly attributable to resource level; we read the pattern as also consistent with structural distance from the typological space the model has internalized.

\textbf{Within-language register as a predicted failure axis.} Informal and code-switched text is a well-documented weak point for NLP systems, including for languages with abundant data \citep{sitaram2020surveycodeswitchedspeechlanguage,aji-etal-2022-one}. We predict that within high-resource pairs such as English--Mandarin, models will perform worse on informal-register items than on formal ones even though informal web text is abundant. Resource scarcity would not explain such a gap \emph{within} an abundantly resourced pair; (M2)-mismatch would: informal usage is the region of $\Phi(C, t)$ where pragmatic-load differentiation is highest, hence where rank-distortion under any single-reference $\tau$ is largest.

\subsection{Probing (M2) directly: BETEL as design illustration}

A second wave of benchmarks moves from translation-pair scoring toward direct probing of cultural-pragmatic competence: BLEnD \citep{myung2024blend}, CDEval \citep{wang-etal-2024-cdeval}, and the in-development BETEL, built on the Taiwan Cultural Reference Benchmark \citep{tseng2026taiwanbenchmark}. Their common move: construct probe items where the (M1) gloss is fixed or trivialized so (M2) competence becomes the discriminating variable.

BETEL illustrates the design space the framework calls for. Its inventory spans five item families (contextual-dialogue comprehension, multilingual lexical-pragmatic function probes, semantic-shift segments, non-literal pragmatic tasks, and pragmatic-appropriateness evaluation), each targeting a structurally distinct (M2)-violation route. Three sample items show the design logic. A \emph{contextual-dialogue} item involving a Tao Indigenous festival tests whether the model recovers the pragmatic significance of an addressee's repeated pauses, which signal an unspoken cross-cultural-knowledge gap that no centroid mapping of the surface utterances exposes. A \emph{multilingual lexical-pragmatic} item presents a Taiwanese Hokkien expression with five candidate situational contexts, probing whether the model recovers the \emph{situational distribution} of the expression: the geometry of $\Phi_{\mathrm{nan}}(C)$, directly testing (M2) on the source-side sub-cloud structure. A \emph{semantic-shift} item targets the diachronic axis of \S\ref{sec:diachrony} by asking the model to reason across historically stratified senses of \emph{q\'{i}ng}.

BETEL items here illustrate the design space; systematic evaluation is forthcoming. The point is that the items show the design space to be \emph{operationally precise}.

\subsection{Summary}

The evidence in hand (the script-vs-family dissociation in FLORES-200, the family-level ordering of LOBSTER scores, and the design space of cultural-pragmatic probes) is consistent with the framework's predictions, and the within-language register gap is a further prediction to test. Two things are missing. \emph{First}, a measurement vocabulary that turns each behavioral signature into a falsifiable, decomposable prediction; \S\ref{sec:metrics} supplies this. \emph{Second}, evidence at the mechanistic level that a regularity the model encodes need not be usable by intervention; \S\ref{sec:rig} offers a first case study. Without such evidence, the behavioral findings remain open to a familiar deflection (``more data will fix it''), which the present evidence does not settle.

\section{The representation--intervention gap}\label{sec:rig}

If unalignability is a real structural property, it should leave a
signature \textbf{inside the model}, not only on its outputs. We argue that such a signature can be looked for, and report a first case study probing for it, in the form of what we call the \textbf{Representation--Intervention Gap (RIG)}: a systematic dissociation between what a model demonstrably \emph{encodes} in its activations and what can be \emph{causally accessed} through principled, locality-based intervention.

\subsection{The probe--causal disjunction}

High probe accuracy demonstrates \emph{decodability}, not \emph{use} \citep{belinkov2022probing,belinkov-glass-2019-analysis,ravichander-etal-2021-probing}. Causal interpretability \citep{geiger2024finding,wang2023interpretabilitywild} formalizes this: $R$ is causally responsible for $b$ only if interventions on $R$ change $b$.

Let $\mathcal{M}$ be a multilingual LM, $C$ a target structure, $T$ a downstream task. Let $\pi(\mathcal{M}, C)$ be probe-decoding accuracy and $\iota(\mathcal{M}, C, T)$ the intervention effect. Define:
\begin{equation}\label{eq:rig}
\mathrm{RIG}(C, T;\mathcal{M}) \;:=\; \widehat\pi(\mathcal{M}, C) \;-\; \widehat\iota(\mathcal{M}, C, T) \;\in\; [-1, 1],
\end{equation}
where $\widehat\pi \in [0,1]$ is probe-decoding accuracy above chance and
$\widehat\iota \in [0,1]$ is the normalized intervention effect on task $T$.
A large positive $\mathrm{RIG}$ (high decodability, vanishing intervention effect) indicates that $C$ is \emph{visible but inoperable}: present in the geometry, but not aligned with the model's modular causal structure. We hypothesize that culturally loaded and typologically specific structures characteristically yield large $\mathrm{RIG}$; the case study below is a first test of this hypothesis, not a confirmation of it.

\subsection{Case study: Yami}

We illustrate RIG with a controlled case study on Yami (also Tao;
ISO~639-3: \texttt{tao}), an endangered Austronesian language spoken on
Orchid Island, Taiwan, whose Tao ethnic population numbers approximately
$5{,}000$ \citep{cip2024population}. The parallel Yami--Mandarin corpus
has $31{,}255$ parallel pairs, of which $24{,}526$ ($335$K Yami tokens)
are used for training. Nine multilingual LMs are evaluated, spanning five
model series (NLLB-200 at 3.3B; decoder-only models 4--32B): Qwen3
\citep{yangQwen3TechnicalReport2025}, Gemma~3 PT
\citep{teamGemma3Technical2025a}, NLLB-200 3.3B encoder--decoder
\citep{nllbteam2022languageleftbehindscaling}, SeaLLMs-v3
\citep{zhangSeaLLMs3Open2024}, and SEA-LION-v4 \citep{sea_lion_2024}
(full lineup in Appendix~\ref{app:design}, Table~\ref{tab:models}). The setup tests whether the \emph{Language Activation
Probability Entropy} (LAPE) method
\citep{tang2024languagespecificneurons}, which identifies
``language-specific neurons'' by their differential activation profile
across languages, can be leveraged to improve low-resource translation.
Each model is fine-tuned under three regimes---Standard LoRA, LAPE-masked
LoRA (MLP LoRA updates gradient-masked to LAPE-selected neurons, at most
$1\%$ of MLP neurons), and random-mask LoRA matched on sparsity---with an
identical, unmasked attention LoRA path in every regime, crossed with
five neuron-sourcing conditions: Yami-derived, Filipino (single
Austronesian), Pan-Austronesian (pooled), English (lingua franca), and
an unrelated-language negative control (Russian, Vietnamese, Korean).
Sixteen paired comparisons are evaluated on the common
Yami$\leftrightarrow$Chinese benchmark at the model level ($n\!=\!9$
model-level differences per comparison, each averaging a model's two
translation directions), reporting the mean difference, a 95\% percentile
bootstrap stability interval ($B\!=\!100{,}000$), paired Cohen's $d_z$, and the
number of models sharing the sign of the estimate. With nine models, exact
sign-flip $p$-values cannot clear Holm--Bonferroni correction over the
sixteen comparisons (minimum adjusted $p = 0.0625$), so we read the results
through estimates, intervals, effect sizes, and sign agreement.\footnote{Full design, all
$16$ comparisons ($48$ comparison--metric cells), and the activation-similarity, mask-overlap,
and per-model failure-mode analyses cited below are in
Appendices~\ref{app:design}--\ref{app:atlas}.}

\textbf{$\widehat\pi$: a regularity is encoded.} We proxy $\widehat\pi$
by activation-profile similarity rather than a trained probe. Among the
eleven comparison languages of the main analysis, Yami's three nearest
neighbors by activation-profile cosine similarity (per-layer cosine,
mean across layers) are Austronesian in all nine models
(Appendix~\ref{app:yami-dendrogram}, Table~\ref{tab:yami_neighbors}),
drawn from \{Filipino, Cebuano, Plateau Malagasy, Maori, Indonesian\}.
In Qwen3-4B the four nearest neighbors are Plateau Malagasy ($0.890$),
Cebuano ($0.890$), Filipino ($0.888$), and Maori ($0.856$); Mandarin
($0.770$) and English ($0.755$) are furthest. In eight of the nine
models the entire Austronesian set is closer to Yami than any
non-Austronesian language; the encoder--decoder NLLB-200 is a partial
exception (Filipino, Cebuano, and Plateau Malagasy remain nearest, but
English and Korean rank above Maori and Indonesian). This regularity is
not cleanly genealogical. Across sixteen evaluation languages, including
five low-resource non-Austronesian controls, distance from Yami
correlates with log FineWeb-2 corpus size, a public proxy for pretraining exposure (Pearson $r = 0.879$);
three of the controls (Yoruba, Igbo, Hausa) fall within the
Austronesian distance range, Indonesian sits with the high-resource
languages, and an Austronesian-specific residual, though in the
predicted direction, is not robustly established: its significance
depends on how it is tested (residual Welch's $t = -2.01$, $p = 0.100$)
\citep{lian2026dissertation}. What the activations encode,
then, is a measurable regularity along which Yami groups with
other low-resource and Austronesian languages, strongly associated with
corpus scale.

\textbf{$\widehat\iota$: no demonstrated advantage over random masks.} Two convergent
findings, in order of evidential weight. First and most diagnostically,
LAPE-masked LoRA shows no demonstrated advantage over random-mask
LoRA matched on sparsity in any paired comparison: on BLEU the paired
Cohen's $d_z \in [-0.18, +0.17]$, every stability interval includes zero,
and only 3--7 of the nine models share the sign of the estimate. The
cleanest test is Exp 1, trained on the benchmark pair
($\Delta = +0.29$ BLEU, interval $[-0.82, 1.24]$, $d_z = 0.17$); in the cross-pair conditions
(Exp 2a, 2b, 3), Yami$\to$Chinese BLEU is $0.000$ for LAPE in all nine
models and for Random in most (all nine in Exp 2b, eight in Exp 2a, six
in Exp 3), so for those models half of the model-level difference is a
floor-versus-floor tie. A null result with nine models does not show
equivalence.
Second, LAPE-masked LoRA underperforms unconstrained Standard LoRA on
every condition ($d_z \in [-2.64, -1.45]$ on BLEU; every stability
interval lies below zero; 8 or 9 of the nine models agree in sign). In Exp 1,
both results also hold separately on short entries and on sentences
(Appendix~\ref{app:bootstrap}).
Neuron identity, as selected by LAPE, shows no demonstrated benefit
beyond what an arbitrary sparsity-matched subset provides. Read against
the encoded regularity above, this is the RIG signature: a regularity
visible in the activations, not usable by a principled
localization-based intervention.

\textbf{The genealogical inversion.} A subsidiary observation: the
proxy-language masks (Filipino, Pan-Austronesian, English)
\emph{underperform} the unrelated-language negative control
($d_z = -1.92, -1.30, -1.47$ on BLEU; stability intervals below zero;
9, 8, and 8 of the nine models agree in sign). Mask identity is not the
only difference: the proxy-language conditions train on Yami--Filipino (Exp 2a, 2b) or
Yami--English (Exp 3), while the negative control trains on the
Yami--Chinese benchmark pair; the Filipino and English sides of the
cross-pair data are machine (pivot) translations of the Chinese side; and
the cross-pair LAPE conditions collapse to $0.000$ BLEU on
Yami$\to$Chinese in all nine models. One
alternative, damage to Mandarin processing through overlap with the
Mandarin mask, is weighed against by direct mask-Jaccard measurement:
the proxy-language masks overlap the LAPE-identified Mandarin mask
\emph{less} than the negative-control mask does
(Appendix~\ref{app:cmn-overlap}). The negative-control languages are
also unrelated to Yami but not to Chinese: Korean and Vietnamese carry
Sino-Korean and Sino-Vietnamese lexical strata, which may favor the
Yami--Chinese benchmark. Because mask identity, training pair,
pivot-translation quality, and contact with Chinese are confounded and
no ablation separating them was run, we
treat the ordering as descriptive and do not rest the RIG claim on it.

\textbf{Scope.} The dissociation we document is between an
activation-level regularity (Yami grouping with other low-resource and
Austronesian languages) and the operability of \emph{LAPE-style
language-specific neuron localization} as the intervention vehicle. We do not claim every locality-based intervention
is doomed; we claim that a widely used principled neuron-attribution method,
tested across nine models, five model series, and 3.3--32B
parameters, does not demonstrate a benefit from the regularity present
in the models' activations. Because every regime also trains an unmasked
attention LoRA path, attention adaptation may carry much of the
translation gain and narrow any LAPE-versus-random difference; the
comparison isolates the added value of MLP-neuron selection, not a fully
neuron-restricted adapter. One candidate explanation of the gap,
registered in the dissertation before the experiments and neither
confirmed nor ruled out by them, is that LAPE's entropy criterion
selects neurons that mark which language is present rather than neurons
whose adaptation improves translation. Further limits: the $1\%$
candidate rate is inherited from \citet{tang2024languagespecificneurons}
and was not varied in training; conditions share an early-stopping rule
rather than a matched compute budget; and for NLLB-200, identification
and masking are decoder-side, while the Austronesian signal is stronger
in its frozen encoder (mean silhouette $0.195$ vs.\ $0.065$). Whether richer schemes---Distributed Alignment
Search \citep{geiger2024finding}, sparse-autoencoder feature
decomposition \citep{huben2024sparse}, attention-head editing---close
the gap is an open empirical question, and exactly the program a
position-paper-stage diagnosis is meant to motivate.

\subsection{What the gap tells us}

\textbf{Encoding does not entail operability.} The architecture--computation distinction familiar from neuroscience \citep{10.1016/j.cogsys.2024.101244} applies in sharpened form: a structure can be present in the \emph{coordinate system} of activations without being present in the \emph{modular causal structure} the model uses to generate behavior. In Marr's terms, decodability is an \emph{implementational}-level claim; intervention effect is an \emph{algorithmic}-level claim, and our hypothesis is that on culturally loaded structures they characteristically come apart.

\textbf{Encoding and use are separate questions.} The models in our case study include NLLB-200, one of the most data-rich multilingual translation systems available, and decoder-only models up to 32B parameters. The case study shows that they encode a regularity relating Yami to other languages, largely tracking corpus scale, and that the intervention tested here does not turn it into a translation benefit. Whether more data would change either side is an open question. Benchmarks scoring only the surface gloss, meanwhile, risk rewarding behavior that bypasses the very structures whose preservation we should be measuring.

\section{Operationalizing unalignability}\label{sec:metrics}

The argument of \S\S\ref{sec:frame}--\ref{sec:rig} needs to bottom out in measurements benchmark designers can reproduce. We propose four metrics, organized as a stack from outermost behavioral surface to innermost causal substrate. Unalignability is not a single quantity but a \emph{profile}; conflating its dimensions destroys exactly the diagnostic power we want to recover.

\paragraph{Metric 1: Cycle-Consistency (CC).}
For concept $C$ with culturally grounded probe contexts $\{x_1, \ldots, x_N\}$, translator $T$, and embedder $\Phi$:
\begin{equation}\label{eq:cc}
\mathrm{CC}(C) \;=\; 1 - \rho_{\mathrm{Spearman}}\!\left(\mathbf{d}_{\mathrm{src}},\,\mathbf{d}_{\mathrm{cyc}}\right),
\end{equation}
where $\mathbf{d}_{\mathrm{src}}$ and $\mathbf{d}_{\mathrm{cyc}}$ are pairwise embedding distances before and after the round-trip $L_A \to L_B \to L_A$. CC tests whether the \emph{relative} manifold structure of the probe set survives translation. This departs from classical backtranslation: BLEU on cycling rewards lexical recovery; CC measures preservation of \emph{contextual contrasts}: whether the relations among usage contexts survive a round-trip translation.

\paragraph{Metric 2: Pragmatic-Load Disagreement (PLD).}
Given a probe item $(C, X, \{R_1, R_2, R_3\})$ with three native-speaker-annotated continuations, compare the model's selection $\hat{a}_M$ against the modal native-speaker selection $a_H^{\star}$, adjusting for inter-annotator agreement (IAA):
\begin{equation}\label{eq:pld}
\mathrm{PLD}(C) \;=\; \mathrm{Disagree}(\hat{a}_M, a_H^{\star}) \;-\; (1 - \mathrm{IAA}_H).
\end{equation}
PLD isolates pragmatic competence in a controlled multiple-choice setting, robust to surface translation quality. It evaluates whether the model picks the right social register.

\paragraph{Metric 3: Manifold-Curvature Mismatch (MCM).}
For $C$ and its putative equivalent $C'$ with embeddings $\mathcal{E}_A, \mathcal{E}_B$:
\begin{equation}\label{eq:mcm}
\mathrm{MCM}(C) \;=\; \big(\,\Delta_{\dim},\; \Delta_{\rho},\; r_{\mathrm{Proc}}\,\big),
\end{equation}
with $\Delta_{\dim}$ the TwoNN intrinsic dimensionality difference \citep{Facco2017EstimatingTI}, $\Delta_\rho$ mean local density difference, and $r_{\mathrm{Proc}}$ the Procrustes residual. The tuple functions as a \emph{geometric fingerprint} (Appendix~\ref{app:figure2}, Figure~\ref{fig:MCM}), capturing whether the two concepts occupy comparable regions of embedding space.

\paragraph{Metric 4: RIG.}
Restated from \S\ref{sec:rig}: $\mathrm{RIG}(C,T;\mathcal{M}) = \widehat\pi - \widehat\iota \in [-1,1]$, where $\widehat\iota$ uses activation patching \citep{wang2023interpretabilitywild} at minimum and, ideally, Distributed Alignment Search \citep{geiger2024finding}.

\paragraph{The unalignability profile.}
We propose reporting $\langle \mathrm{CC}, \mathrm{PLD}, \mathrm{MCM}, \mathrm{RIG}\rangle$ per concept. Diagnostic patterns are shown in Table~\ref{table1}.

\begin{table}[h]
\centering
\caption{Diagnostic patterns of the four-metric unalignability profile.}
\label{table1}
\small
\begin{tabular}{@{}p{0.35\linewidth}p{0.58\linewidth}@{}}
\toprule
\textbf{Profile} & \textbf{Diagnosis} \\
\midrule
Low CC, low PLD, low MCM, low RIG & Genuinely alignable; standard translation suffices. \\
High CC, low PLD, low RIG & Lexical drift but pragmatic competence intact (\emph{scaling-tractable}). \\
Low CC, high PLD & Pragmatic load lost. \emph{The most common failure of current benchmarks.} \\
High MCM, high RIG & Geometry differs \emph{and} the model cannot recruit the difference. \emph{The hard case.} \\
Any high, RIG $\approx 0$ & Failure is behavioral, not architectural; better data plausibly closes it. \\
\bottomrule
\end{tabular}
\end{table}

A test set is well-constructed not when it produces a clean leaderboard, but when concepts in the high-MCM-high-RIG quadrant are \emph{visible as such} in the evaluation: these metrics are diagnostics for the multidimensional nature of alignment failure, not improvements to evaluation.

\section{Recommendations}\label{sec:rec}

\textbf{For benchmark designers.} We propose replacing scalar rankings with an Unalignability Profile: report $\langle \mathrm{CC},\mathrm{PLD},\mathrm{MCM},\mathrm{RIG} \rangle$ rather than a single score such as BLEU or chrF; stratify by typological feature; include diachronic items; author multiple registered translations per source item, admitting \emph{no faithful gloss} as a possible answer. We advocate a leaderboard that stratifies models by their ability to maintain structural faithfulness (M2) across typologically distant families.

\textbf{For model designers.} We hypothesize that data quantity alone will not remove structural distortion. The second author's dissertation \citep{lian2026dissertation} finds that interventions on language-specific neuron sets show no demonstrated advantage over random-mask controls. Future fine-tuning should include explicit (M2)-conditioned supervision (paraphrase-set targets, pragmatic-context-conditioned loss) rather than additional source-side parallel text alone.

\textbf{For mechanistic-interpretability researchers.} RIG is testable beyond the multilingual case. Construct causal-abstraction probes for any predicted-unalignable concept class and report the (probe-accuracy, intervention-effect) pair.

\section{Discussion and conclusion}\label{sec:disc}

\textbf{Scope and limitations.} The framework is \emph{encoder-relative}: $\Phi_L(C, t)$ is defined relative to a specific $\mathcal{M}$. Two encoders may disagree on $\alpha$-unalignability. We treat this as a feature: unalignability \emph{should} be encoder-relative if it is to diagnose an encoder's alignment behavior.

\textbf{Broader implications.} The pattern we identify, namely representational regularity that is encoded but not usable by intervention, is a candidate diagnostic wherever language models are asked to behave in accordance with structure they internalize without instrumenting: value alignment, calibration of refusals, faithfulness of chain-of-thought, in-context instruction recovery. Cross-cultural unalignability is the linguist's controlled experiment for an alignment problem the field has yet to formalize.

\textbf{Conclusion.} A class of concepts admits no faithful alignment map between languages, by virtue of how their usage clouds differ in pragmatic load, diachronic stratification, and categorial articulation. We propose that the property has a mechanistic signature, the Representation--Intervention Gap, and report a first case study in which an encoded regularity is not usable by a principled intervention. Recognizing this class is the precondition both for benchmarks measuring cultural competence rather than surface translation, and, we argue, for fine-tuning regimes that target structural alignment directly rather than relying on data quantity alone.

\newpage

\bibliographystyle{plainnat}
\bibliography{NeurIPS26}

\newpage
\appendix

\section{Visualization of the cloud-mapping framework}\label{app:figure}

\begin{figure}[h]
  \centering
  \includegraphics[width=0.85\linewidth]{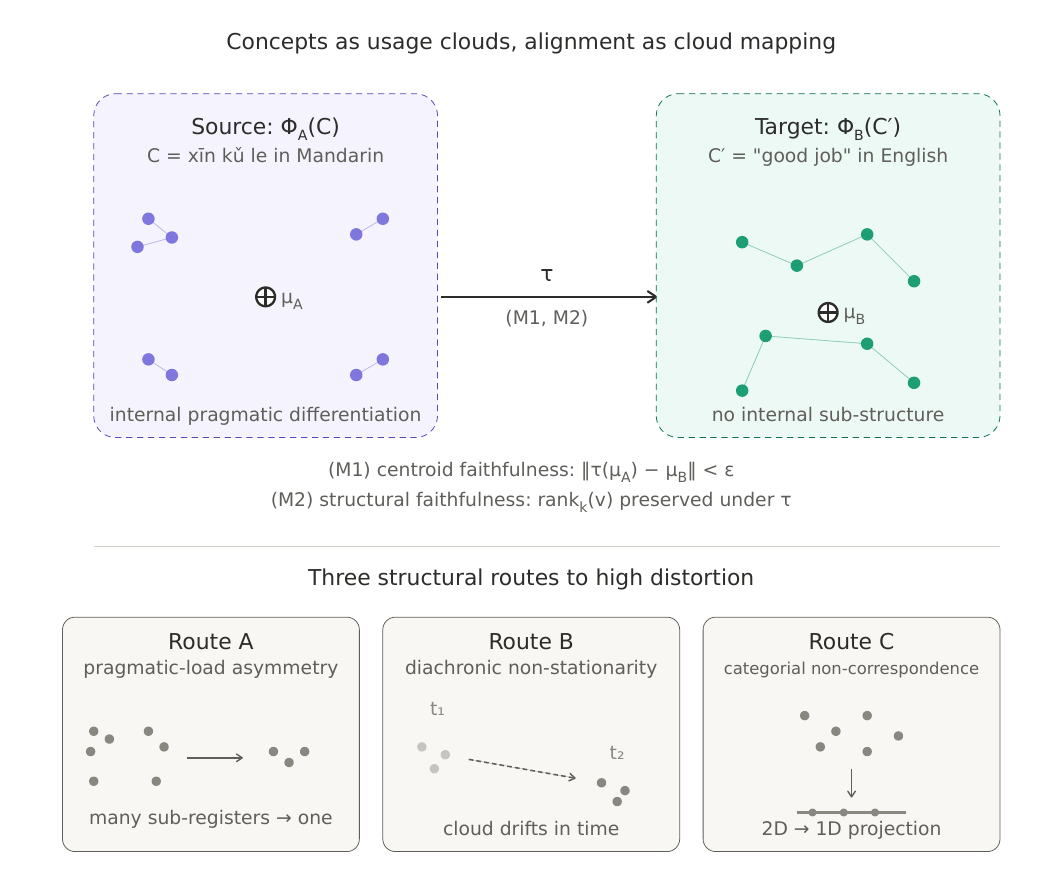}
  \caption{\textbf{Concepts as usage clouds, alignment as cloud mapping} (visualization of the framework introduced in \S\ref{sec:frame}). \textit{Top}: a source cloud $\Phi_A(\textit{x\={\i}n k\v{u} le})$ exhibits internal pragmatic differentiation, with four sub-regions corresponding to superior-to-subordinate, peer-to-peer, service-encounter, and ironic uses; the target cloud $\Phi_B(\textit{good job})$ has no corresponding sub-structure. The alignment map $\tau$ is judged on two conditions: (M1) centroid faithfulness, $\|\tau(\mu_A) - \mu_B\| < \varepsilon$, and (M2) structural faithfulness, preservation of $\mathrm{rank}_k(v)$ under $\tau$. Even when $\tau$ satisfies (M1) by aligning centroids, (M2) is necessarily violated when the target lacks the source's pragmatic dimensionality. \textit{Bottom}: the three structural routes to a large infimum in Definition~3. Route~A (pragmatic-load asymmetry): many sub-registers in the source collapse to one in the target. Route~B (diachronic non-stationarity): the cloud drifts across time windows $t_1 \to t_2$, so no synchronic gloss can simultaneously match both. Route~C (categorial non-correspondence): the target cloud is a projection of the source onto a lower-dimensional subspace.}
  \label{fig:clouds}
\end{figure}

\clearpage
\section{Visualization of the Procrustes residual (\texorpdfstring{$r_{Proc}$}{r\_Proc})}
\label{app:figure2}

\begin{figure}[h]
  \centering
  \includegraphics[width=0.85\linewidth]{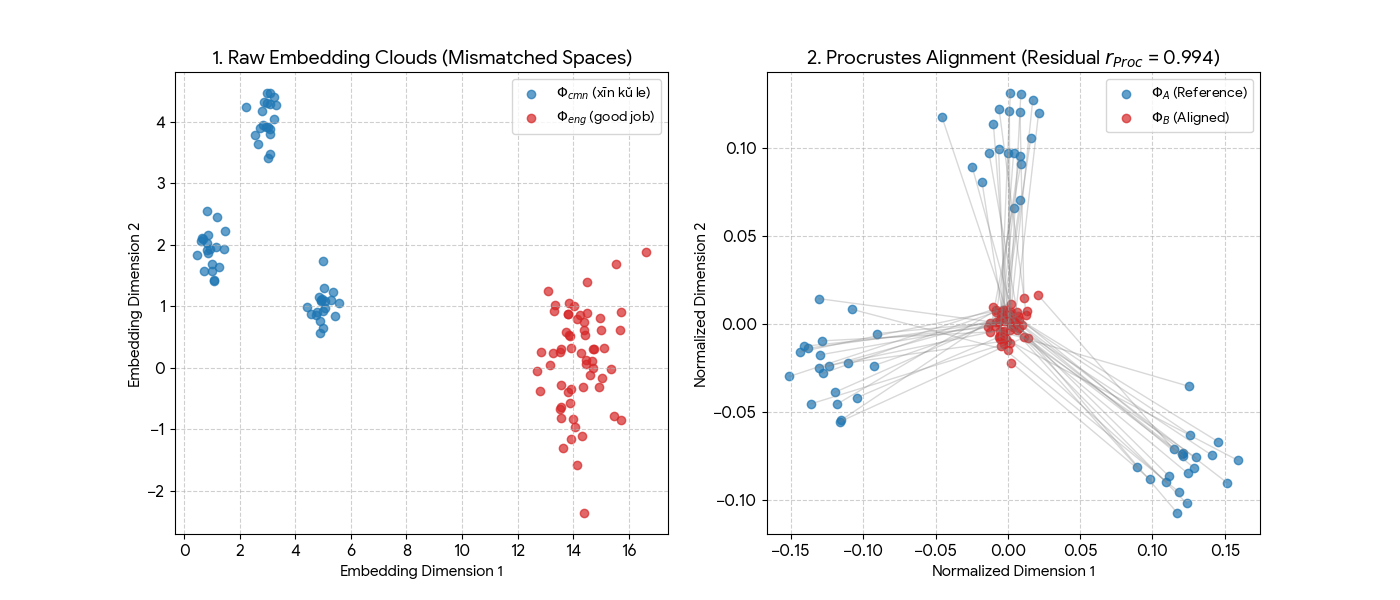}
  \caption{Illustrative example with synthetic clouds: the Procrustes residual $r_{\mathrm{Proc}}$ (Metric~3, Manifold-Curvature Mismatch) for a structured source cloud and a flat target cloud. \textit{Left}: the raw usage clouds $\Phi_{\mathrm{cmn}}(\textit{x\={\i}n k\v{u} le})$ (blue), partitioned into distinct pragmatic regions (superior-to-subordinate, peer-to-peer, service encounter), and $\Phi_{\mathrm{eng}}(\textit{good job})$ (red). \textit{Right}: the clouds after optimal Procrustes alignment (translation, rotation, and scaling); gray lines connect corresponding points. $r_{\mathrm{Proc}}$ is the sum of squared distances between corresponding points after alignment. Even the best linear map cannot stretch the flat target cloud over the sub-clusters of the source cloud, so $r_{\mathrm{Proc}}$ remains high: the internal rank structure and local neighborhood densities are incompatible, an (M2) violation.}
  \label{fig:MCM}
\end{figure}

\clearpage
\section{Experimental design}\label{app:design}

\subsection{Yami corpus and language situation}

Yami (also Tao; ISO 639-3: \texttt{tao}) is an endangered
Austronesian language spoken on Orchid Island (Lanyu), Taiwan, with
a Tao ethnic population of 5{,}045 \citep{cip2024population}; this is a
population figure, not a speaker count. The parallel Yami--Mandarin
corpus, provided by D.~Victoria Rau and Ann Chang, consists of 31{,}255
parallel pairs drawn from Bible translations, a cultural dictionary,
teaching materials, a grammar overview, and an online learning platform,
split into
training (24{,}526 pairs; approximately 335K Yami tokens), validation
(3{,}063), general test (3{,}066), Bible test (500), and a 100-pair
few-shot pool. About half of the material is short entries, defined by
length as at most two Yami words: 11{,}416 of the 24{,}526 training pairs
(46.5\%) and 1{,}700 of the 3{,}566 test pairs (47.7\%). The Yami--Filipino and Yami--English corpora used in
Exp 2a, 2b, and 3 keep the same Yami sentences, with the Filipino and
English sides produced by GPT-5-mini (\texttt{gpt-5-mini-2025-08-07})
pivot translation from the Chinese side; variation in pivot-translation
fidelity may therefore also affect the cross-pair results.
Yami$\leftrightarrow$Chinese is the common evaluation benchmark across all
nine models and all experimental conditions; for cross-pair conditions
(Exp 2a Filipino, Exp 2b Pan-Austronesian, Exp 3 English) the matched
proxy-pair test set is also evaluated. The combined Yami$\leftrightarrow$Chinese
test set (\texttt{general} + \texttt{bible} splits) contains 3{,}566 items.
Held-out status is at the level of complete pairs: no test pair occurs
in the training data, but 708 of the 3{,}566 test pairs reuse a Yami or
Chinese sentence that also occurs in training with a different
translation, so the evaluation is pair-level rather than string-level
held-out.

\subsection{Models}

Nine multilingual language models from five model series are used
(Table~\ref{tab:models}). The LAPE method is applied to decoder MLP
activations for all models.

Decoder-only Qwen3 \citep{yangQwen3TechnicalReport2025} (119 languages,
with explicit Austronesian coverage of Cebuano, Ilocano, Indonesian,
Javanese, Sundanese, Tagalog, and Waray) and Gemma~3 PT
\citep{teamGemma3Technical2025a} (pre-instruction-tuning checkpoints)
span a 4--32B parameter range. SeaLLMs-v3 \citep{zhangSeaLLMs3Open2024}
is SEA-focused and trained with language-specific neuron identification
at the continued-pretraining stage. SEA-LION-v4 \citep{sea_lion_2024}
is continued-pretrained from the instruction-tuned Gemma~3 27B checkpoint
on 500B tokens spanning 11 Southeast Asian languages. NLLB-200
\citep{nllbteam2022languageleftbehindscaling} is the encoder--decoder
translation model (202 languages); LAPE
\citep{tang2024languagespecificneurons} is applied to the decoder side
only. Checkpoint post-training status is not uniform: the Qwen3 runs use
the post-trained (non-base) checkpoints, Gemma~3 uses the pretrained
checkpoints, and SEA-LION-v4 descends from an instruction-tuned
checkpoint. Because Yami is not an NLLB-200 language, NLLB-200 was
extended with a \texttt{tao\_Latn} language token initialized by
cloning the Tagalog embedding.

\begin{table}[h]
\centering\small
\caption{Models in the case study. $^{*}$Extended with a \texttt{tao\_Latn} language token before fine-tuning.}
\label{tab:models}
\begin{tabular}{l l l l}
\toprule
Family & Sizes & Architecture & HuggingFace identifier \\
\midrule
Qwen3       & 4B / 14B / 32B & Decoder-only    & \texttt{Qwen/Qwen3-\{4B,14B,32B\}} \\
Gemma 3 PT  & 4B / 12B / 27B & Decoder-only    & \texttt{google/gemma-3-\{4b,12b,27b\}-pt} \\
NLLB-200    & 3.3B           & Enc--dec        & \texttt{facebook/nllb-200-3.3B}$^{*}$ \\
SeaLLMs-v3  & 7B             & Decoder-only    & \texttt{SeaLLMs/SeaLLMs-v3-7B} \\
SEA-LION-v4 & 27B            & Decoder-only    & \texttt{aisingapore/Gemma-SEA-LION-v4-27B} \\
\bottomrule
\end{tabular}
\end{table}

\subsection{Conditions and modes}

The experimental grid is five \emph{neuron-sourcing conditions} crossed
with three \emph{fine-tuning modes} (Table~\ref{tab:conditions}). Each
LAPE condition is paired with a random-mask control matched on sparsity
(Exp 0 has no random control because it is itself a negative control),
and with a Standard LoRA baseline trained on the same data.

\begin{table}[h]
\centering\small
\caption{Five neuron-sourcing conditions; LAPE selects the listed
target-set neurons (relative to the listed contrast set) via Language
Activation Probability Entropy. Each condition is fine-tuned on the
listed parallel pair under three modes: Standard LoRA (no mask), LAPE
(gradient-masked LoRA on the LAPE-selected neurons), and Random Mask
(matched-sparsity LoRA on a random neuron subset; absent for Exp 0).
The common evaluation benchmark is Yami$\leftrightarrow$Chinese for all
conditions.}
\label{tab:conditions}
\begin{tabular}{l l l l l}
\toprule
ID & Condition & LAPE target & LAPE contrast & Fine-tune pair \\
\midrule
Exp 0 & Neg.\ Ctrl & \{rus, vie, kor\} & \{cmn, arb, fil, eng\} & Yami--Chinese \\
Exp 1 & Yami & \{tao\} & \{cmn, kor, rus, vie, arb\} & Yami--Chinese \\
Exp 2a & Filipino & \{fil\} & \{cmn, kor, rus, arb, vie, eng\} & Yami--Filipino \\
Exp 2b & Pan-AN & \{fil, ceb, ind, mri, plt\} & \{cmn, kor, rus, arb, vie, eng\} & Yami--Filipino \\
Exp 3 & English & \{eng\} & \{cmn, kor, rus, fil\} & Yami--English \\
\bottomrule
\end{tabular}
\end{table}

\subsection{LoRA and gradient-masking implementation}

Each model is trained on bidirectional data (Yami$\to$X and X$\to$Yami
in a 50/50 ratio). LoRA (rank 16, $\alpha=32$, dropout 0.1) over 4-bit
NF4-quantized base weights with bfloat16 compute is applied to the attention
projections (query, key, value, output) and the MLP projections of every
decoder layer: \texttt{gate\_proj}, \texttt{up\_proj}, and
\texttt{down\_proj} for the decoder-only models, and \texttt{fc1} and
\texttt{fc2} for the NLLB-200 decoder (whose cross-attention also
receives LoRA; its encoder is frozen). Training uses 8-bit AdamW with a
cosine-decayed peak learning rate of $1\!\times\!10^{-4}$, effective
batch size 64, and up to 100 epochs with early stopping on validation
loss (patience 3). Sparse masking (LAPE and Random modes) applies only to
the MLP LoRA path; the attention LoRA path is unmasked and identically
configured in every condition. Masking is implemented as gradient masking
on LoRA weights: backward hooks zero gradients for non-target neurons on
the rows of \texttt{lora\_B} for \texttt{gate\_proj}/\texttt{up\_proj}
(or \texttt{fc1}; neuron creation) and on the columns of
\texttt{lora\_A} for \texttt{down\_proj} (or \texttt{fc2}; neuron
consumption).
LoRA parameters for non-target neurons are also explicitly initialized to
zero before training. Sparsity rates (number of LAPE-target neurons
divided by total MLP-intermediate neurons) for the Exp 1 (Yami) masks
range from 0.63\% (NLLB-200, decoder only) to 0.94\% (Qwen3-4B),
averaging approximately 0.9\% across the nine models; across all masks
they range from about 0.4\% to 1.0\%.

\subsection{Inference and evaluation}

Inference uses the HuggingFace \texttt{transformers} library. Decoder-only models load
via \texttt{AutoModelForCausalLM} + \texttt{AutoTokenizer} and generate with
five-shot prompting (5 in-context examples sampled from a held-out
few-shot pool; greedy decoding, \texttt{bfloat16}). NLLB-200 loads via
\texttt{AutoModelForSeq2SeqLM} + \texttt{NllbTokenizer} and translates
source text directly with the appropriate \texttt{forced\_bos\_token\_id}
for the target language; no few-shot prompting is used for NLLB. Translations are evaluated on the
Yami$\leftrightarrow$Chinese common benchmark (3{,}566 items, combining
\texttt{general} and \texttt{bible} splits) using BLEU, chrF++, and GLEU
via the HuggingFace \texttt{evaluate} library, on the standard $[0,100]$
scale. Cross-pair conditions (Exp 2a, 2b, 3) additionally evaluate on
the matched proxy-pair test set. There, LAPE still underperforms
Standard LoRA ($d_z$ from $-2.09$ to $-1.73$; Holm-adjusted
$p \le 0.039$ in this seven-comparison family, whose floor is $0.0273$)
and shows no advantage over Random (Holm-adjusted $p = 1.000$), so the
LAPE systems do function on their training pair while the
common-benchmark conclusion holds.

\subsection{Statistical methodology}

The main analysis is at the model level. For each comparison, each
model's two direction-level paired differences on the common
Yami$\leftrightarrow$Chinese benchmark are averaged, giving $n=9$
model-level differences; the two directions are not treated as
independent because they share one trained adapter. A percentile
bootstrap over these differences ($B=100{,}000$) gives a 95\% stability
interval, and exhaustive sign-flip enumeration ($2^9=512$ assignments)
gives the $p$-value, corrected via Holm--Bonferroni
\citep{holm1979simple} over the sixteen comparisons within each metric
(Appendix~\ref{app:bootstrap}). The smallest attainable two-sided raw
$p$-value is $2/2^9$, so the smallest Holm-adjusted value is
$16 \times 2/512 = 0.0625$ and no comparison can clear $\alpha=0.05$;
we therefore report estimates, stability intervals, effect sizes, and the
number of models sharing the sign of each estimate. Effect sizes are reported as
paired-samples Cohen's $d_z = \overline{\Delta}/\mathrm{sd}(\Delta)$
\citep{lakens2013calculating}. A sentence-level paired bootstrap
\citep{koehn2004statistical}, computed per $(\text{model}, \text{direction})$
slot, is reported as a supplementary robustness check and agrees with the
model-level direction in 78.6\% of the 864 slot$\times$metric cells.

\subsection{Training summary}

Conditions are compared under a common early-stopping rule, not a
matched compute budget. Under that rule the masked Exp 1 runs take
1.17--1.50 times as many optimization steps as Standard LoRA in the
decoder-only models (median 1.45 times the wall-clock time across all
nine models) and still
score lower; performance at a fixed compute budget was not measured.
Table~\ref{tab:training_summary} lists final training and validation
losses, steps, and wall-clock time for every LAPE and Random run, for the
matched Yami--Chinese Standard LoRA baseline, and averaged over the three
Standard LoRA baselines.

\begingroup
\small
\setlength{\tabcolsep}{4pt}
\renewcommand{\arraystretch}{0.95}
\begin{longtable}{@{}ll rr r r@{}}
\caption{Training summary across models and experimental conditions. Train/validation loss are final values; time is wall-clock hours. The Standard LoRA row averages the Yami--Chinese, Yami--English, and Yami--Filipino baselines; Standard LoRA (Yami--Chinese) is the matched baseline for Exp~1. LAPE/Random entries are single runs.}\label{tab:training_summary}\\
\toprule
Model & Condition & Train Loss & Val.\ Loss & Steps & Time (h) \\
\midrule
\endfirsthead
\multicolumn{6}{c}{\textit{Table~\ref{tab:training_summary} (continued)}}\\
\toprule
Model & Condition & Train Loss & Val.\ Loss & Steps & Time (h) \\
\midrule
\endhead
    Qwen3-4B & Standard LoRA & 0.628 & 2.048 & 5369 & 1.3 \\
     & Standard LoRA (Yami--Chinese) & 0.491 & 1.947 & 5369 & 1.3 \\
     & Exp 0 (Neg. Ctrl) [LAPE] & 0.981 & 2.061 & 8437 & 2.0 \\
     & Exp 1 (Yami) [LAPE] & 1.076 & 2.040 & 7670 & 1.8 \\
     & Exp 1 (Yami) [Random] & 0.977 & 2.071 & 8437 & 2.0 \\
     & Exp 2a (Filipino) [LAPE] & 1.347 & 2.072 & 7670 & 2.2 \\
     & Exp 2a (Filipino) [Random] & 1.294 & 2.119 & 8437 & 2.4 \\
     & Exp 2b (Pan-AN) [LAPE] & 1.338 & 2.086 & 7670 & 2.2 \\
     & Exp 2b (Pan-AN) [Random] & 1.355 & 2.089 & 7670 & 2.2 \\
     & Exp 3 (English) [LAPE] & 1.182 & 2.100 & 7670 & 1.8 \\
     & Exp 3 (English) [Random] & 1.076 & 2.161 & 8437 & 2.0 \\
\midrule
    Qwen3-14B & Standard LoRA & 0.469 & 1.853 & 4602 & 2.5 \\
     & Standard LoRA (Yami--Chinese) & 0.414 & 1.687 & 4602 & 2.4 \\
     & Exp 0 (Neg. Ctrl) [LAPE] & 0.852 & 1.807 & 6903 & 3.8 \\
     & Exp 1 (Yami) [LAPE] & 0.793 & 1.822 & 6903 & 3.8 \\
     & Exp 1 (Yami) [Random] & 0.835 & 1.803 & 6903 & 3.8 \\
     & Exp 2a (Filipino) [LAPE] & 0.967 & 1.911 & 6903 & 4.4 \\
     & Exp 2a (Filipino) [Random] & 1.006 & 1.902 & 6903 & 4.4 \\
     & Exp 2b (Pan-AN) [LAPE] & 0.961 & 1.901 & 6903 & 4.4 \\
     & Exp 2b (Pan-AN) [Random] & 0.998 & 1.916 & 6903 & 4.5 \\
     & Exp 3 (English) [LAPE] & 0.929 & 1.942 & 6903 & 3.7 \\
     & Exp 3 (English) [Random] & 0.946 & 1.962 & 6903 & 3.7 \\
\midrule
    Qwen3-32B & Standard LoRA & 0.343 & 1.833 & 4602 & 6.0 \\
     & Standard LoRA (Yami--Chinese) & 0.294 & 1.682 & 4602 & 5.9 \\
     & Exp 0 (Neg. Ctrl) [LAPE] & 0.686 & 1.748 & 6903 & 9.6 \\
     & Exp 1 (Yami) [LAPE] & 0.651 & 1.783 & 6903 & 9.7 \\
     & Exp 1 (Yami) [Random] & 0.753 & 1.702 & 6136 & 13.5 \\
     & Exp 2a (Filipino) [LAPE] & 0.893 & 1.806 & 6136 & 9.4 \\
     & Exp 2a (Filipino) [Random] & 0.944 & 1.801 & 6136 & 9.3 \\
     & Exp 2b (Pan-AN) [LAPE] & 0.891 & 1.826 & 6136 & 9.4 \\
     & Exp 2b (Pan-AN) [Random] & 0.932 & 1.824 & 6136 & 9.3 \\
     & Exp 3 (English) [LAPE] & 0.765 & 1.925 & 6903 & 9.4 \\
     & Exp 3 (English) [Random] & 0.761 & 1.933 & 6903 & 9.4 \\
\midrule
    Gemma-3-4B & Standard LoRA & 0.626 & 2.294 & 5369 & 6.3 \\
     & Standard LoRA (Yami--Chinese) & 0.551 & 2.294 & 5369 & 6.3 \\
     & Exp 0 (Neg. Ctrl) [LAPE] & 1.277 & 2.224 & 7670 & 10.0 \\
     & Exp 1 (Yami) [LAPE] & 1.190 & 2.220 & 7670 & 9.9 \\
     & Exp 1 (Yami) [Random] & 1.143 & 2.291 & 8437 & 11.1 \\
     & Exp 2a (Filipino) [LAPE] & 1.343 & 2.157 & 7670 & 10.3 \\
     & Exp 2a (Filipino) [Random] & 1.332 & 2.155 & 8437 & 11.3 \\
     & Exp 2b (Pan-AN) [LAPE] & 1.354 & 2.150 & 7670 & 10.3 \\
     & Exp 2b (Pan-AN) [Random] & 1.419 & 2.175 & 7670 & 10.2 \\
     & Exp 3 (English) [LAPE] & 1.400 & 2.150 & 6903 & 9.0 \\
     & Exp 3 (English) [Random] & 1.227 & 2.215 & 8437 & 10.9 \\
\midrule
    Gemma-3-12B & Standard LoRA & 0.436 & 2.043 & 4602 & 2.5 \\
     & Standard LoRA (Yami--Chinese) & 0.414 & 2.027 & 4602 & 2.4 \\
     & Exp 0 (Neg. Ctrl) [LAPE] & 0.851 & 2.100 & 6136 & 3.4 \\
     & Exp 1 (Yami) [LAPE] & 0.759 & 2.099 & 6136 & 3.4 \\
     & Exp 1 (Yami) [Random] & 0.837 & 2.104 & 6136 & 3.5 \\
     & Exp 2a (Filipino) [LAPE] & 0.849 & 1.998 & 6136 & 3.8 \\
     & Exp 2a (Filipino) [Random] & 0.924 & 1.988 & 6136 & 3.8 \\
     & Exp 2b (Pan-AN) [LAPE] & 0.866 & 1.940 & 6136 & 3.8 \\
     & Exp 2b (Pan-AN) [Random] & 0.913 & 2.002 & 6136 & 3.8 \\
     & Exp 3 (English) [LAPE] & 0.954 & 2.017 & 6136 & 3.4 \\
     & Exp 3 (English) [Random] & 0.925 & 2.002 & 6136 & 3.5 \\
\midrule
    Gemma-3-27B & Standard LoRA & 0.356 & 1.938 & 4602 & 5.7 \\
     & Standard LoRA (Yami--Chinese) & 0.347 & 1.837 & 4602 & 5.6 \\
     & Exp 0 (Neg. Ctrl) [LAPE] & 0.672 & 1.827 & 5369 & 7.3 \\
     & Exp 1 (Yami) [LAPE] & 0.614 & 1.859 & 5369 & 7.3 \\
     & Exp 1 (Yami) [Random] & 0.697 & 1.812 & 5369 & 7.2 \\
     & Exp 2a (Filipino) [LAPE] & 0.773 & 1.830 & 5369 & 7.7 \\
     & Exp 2a (Filipino) [Random] & 0.682 & 1.976 & 6136 & 8.8 \\
     & Exp 2b (Pan-AN) [LAPE] & 0.598 & 1.991 & 6136 & 8.8 \\
     & Exp 2b (Pan-AN) [Random] & 0.685 & 1.917 & 6136 & 8.9 \\
     & Exp 3 (English) [LAPE] & 0.648 & 1.974 & 6136 & 8.3 \\
     & Exp 3 (English) [Random] & 0.691 & 1.992 & 6136 & 8.2 \\
\midrule
    NLLB-200-3.3B & Standard LoRA & 1.323 & 2.121 & 26845 & 2.8 \\
     & Standard LoRA (Yami--Chinese) & 1.048 & 2.054 & 36816 & 3.7 \\
     & Exp 0 (Neg. Ctrl) [LAPE] & 2.601 & 2.092 & 49088 & 6.9 \\
     & Exp 1 (Yami) [LAPE] & 1.503 & 2.119 & 36816 & 3.9 \\
     & Exp 1 (Yami) [Random] & 10.958 & 2.076 & 55991 & 28.6 \\
     & Exp 2a (Filipino) [LAPE] & 1.591 & 2.274 & 35282 & 3.9 \\
     & Exp 2a (Filipino) [Random] & 14.708 & 2.277 & 32214 & 16.6 \\
     & Exp 2b (Pan-AN) [LAPE] & 3.197 & 2.272 & 35282 & 5.0 \\
     & Exp 2b (Pan-AN) [Random] & 12.744 & 2.263 & 42952 & 22.2 \\
     & Exp 3 (English) [LAPE] & 3.052 & 2.070 & 28379 & 4.0 \\
     & Exp 3 (English) [Random] & 1.567 & 2.079 & 25311 & 2.7 \\
\midrule
    SeaLLMs-v3-7B & Standard LoRA & 0.466 & 2.010 & 4857 & 1.5 \\
     & Standard LoRA (Yami--Chinese) & 0.432 & 1.837 & 4602 & 1.4 \\
     & Exp 0 (Neg. Ctrl) [LAPE] & 0.950 & 1.963 & 7670 & 2.5 \\
     & Exp 1 (Yami) [LAPE] & 1.049 & 1.925 & 6903 & 2.2 \\
     & Exp 1 (Yami) [Random] & 0.858 & 2.001 & 8437 & 2.7 \\
     & Exp 2a (Filipino) [LAPE] & 1.217 & 2.076 & 7670 & 2.8 \\
     & Exp 2a (Filipino) [Random] & 1.276 & 2.070 & 7670 & 2.8 \\
     & Exp 2b (Pan-AN) [LAPE] & 1.211 & 2.092 & 7670 & 2.8 \\
     & Exp 2b (Pan-AN) [Random] & 1.281 & 2.080 & 7670 & 2.8 \\
     & Exp 3 (English) [LAPE] & 1.058 & 2.026 & 7670 & 2.3 \\
     & Exp 3 (English) [Random] & 1.044 & 2.046 & 7670 & 2.4 \\
\midrule
    SEA-LION-v4-27B & Standard LoRA & 0.356 & 2.067 & 4602 & 5.7 \\
     & Standard LoRA (Yami--Chinese) & 0.353 & 2.013 & 4602 & 5.6 \\
     & Exp 0 (Neg. Ctrl) [LAPE] & 0.717 & 2.054 & 5369 & 7.3 \\
     & Exp 1 (Yami) [LAPE] & 0.662 & 2.001 & 5369 & 7.2 \\
     & Exp 1 (Yami) [Random] & 0.740 & 1.990 & 5369 & 7.3 \\
     & Exp 2a (Filipino) [LAPE] & 0.797 & 1.900 & 5369 & 7.6 \\
     & Exp 2a (Filipino) [Random] & 0.888 & 1.869 & 5369 & 7.7 \\
     & Exp 2b (Pan-AN) [LAPE] & 0.809 & 1.927 & 5369 & 7.8 \\
     & Exp 2b (Pan-AN) [Random] & 0.709 & 2.033 & 6136 & 9.1 \\
     & Exp 3 (English) [LAPE] & 0.858 & 1.988 & 5369 & 7.2 \\
     & Exp 3 (English) [Random] & 0.645 & 2.140 & 6136 & 8.2 \\
\bottomrule
\end{longtable}
\endgroup

\clearpage
\section{Bootstrap results}\label{app:bootstrap}

\begin{figure}[h]
  \centering
  \includegraphics[width=0.95\linewidth]{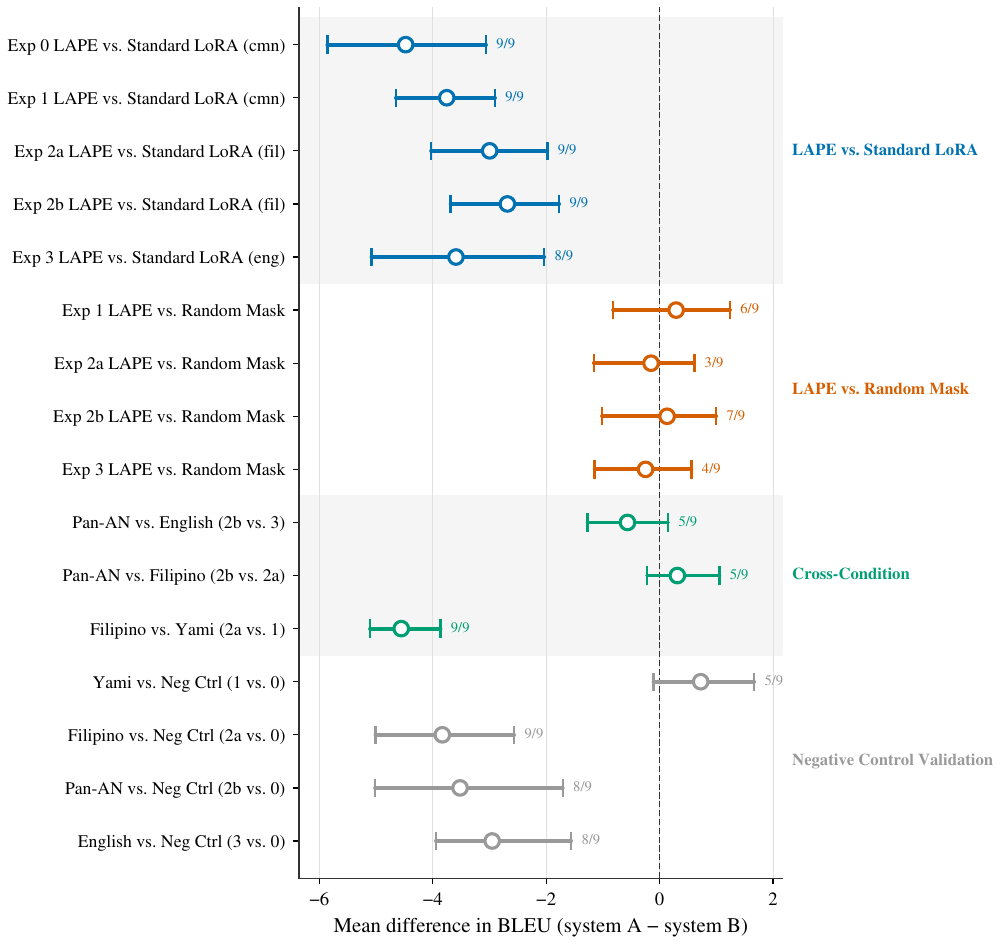}
  \caption{Forest plot of model-level mean paired BLEU differences
  (system A $-$ system B) with 95\% percentile stability intervals for all
  16 comparisons on the Yami$\leftrightarrow$Chinese common benchmark.
  Labels give the number of the nine models sharing the sign of the
  estimate. Comparisons are grouped by family (LAPE-vs-Standard,
  LAPE-vs-Random, cross-condition, negative-control validation). The
  LAPE-vs-Random intervals all include zero: no demonstrated LAPE
  advantage, which is not evidence of equivalence.}
  \label{fig:forest}
\end{figure}

The 16 paired comparisons fall into four families (5 LAPE-vs-Standard,
4 LAPE-vs-Random, 3 cross-condition research-question comparisons,
and 4 negative-control validations). Model-level estimates, stability
intervals, effect sizes, and sign-agreement counts are in
Table~\ref{tab:bootstrap-summary}; the sign-agreement summary across the
three metrics is in Table~\ref{tab:significance-matrix}, and
Figure~\ref{fig:forest} plots the BLEU estimates. Holm-corrected
$p$-values are floor-limited by design (minimum $0.0625$;
Appendix~\ref{app:design}) and are not used as decision criteria.

\begingroup
\small
\setlength{\tabcolsep}{4pt}
\renewcommand{\arraystretch}{0.95}
\begin{longtable}{@{}lllccr@{}}
\caption{Main model-level paired-analysis summary for the 16 pre-specified comparisons on the common Yami$\leftrightarrow$Chinese benchmark. $\Delta$: model-level mean difference $\overline{A-B}$ (BLEU and GLEU on $[0,100]$ scale; chrF++ native). 95\% stability interval: percentile interval over resampled model-level differences. Agree: number of trained models whose difference shares the sign of the overall estimate. $d_z$: paired-samples Cohen's $d_z$ (mean paired difference divided by SD of paired differences).}\label{tab:bootstrap-summary}\\
\toprule
Comparison & Metric & $\Delta$ & 95\% stab.\ int. & $d_z$ & Agree \\
\midrule
\endfirsthead
\multicolumn{6}{c}{\textit{Table~\ref{tab:bootstrap-summary} (continued)}}\\
\toprule
Comparison & Metric & $\Delta$ & 95\% stab.\ int. & $d_z$ & Agree \\
\midrule
\endhead
\multirow{3}{*}{Exp 0 LAPE vs.\ Standard LoRA (cmn)} & BLEU & $-$4.48 & [$-$5.86, $-$3.06] & $-$1.96 & 9/9 \\*
 & chrF++ & $-$5.97 & [$-$7.76, $-$4.14] & $-$2.04 & 9/9 \\*
 & GLEU & $-$3.76 & [$-$4.96, $-$2.51] & $-$1.87 & 9/9 \\
\cmidrule(lr){1-6}
\multirow{3}{*}{Exp 1 LAPE vs.\ Standard LoRA (cmn)} & BLEU & $-$3.75 & [$-$4.65, $-$2.90] & $-$2.64 & 9/9 \\*
 & chrF++ & $-$4.69 & [$-$5.49, $-$3.83] & $-$3.48 & 9/9 \\*
 & GLEU & $-$2.84 & [$-$3.53, $-$2.17] & $-$2.57 & 9/9 \\
\cmidrule(lr){1-6}
\multirow{3}{*}{Exp 2a LAPE vs.\ Standard LoRA (fil)} & BLEU & $-$3.00 & [$-$4.03, $-$1.98] & $-$1.80 & 9/9 \\*
 & chrF++ & $-$4.17 & [$-$5.53, $-$2.91] & $-$1.99 & 9/9 \\*
 & GLEU & $-$2.72 & [$-$3.78, $-$1.77] & $-$1.67 & 9/9 \\
\cmidrule(lr){1-6}
\multirow{3}{*}{Exp 2b LAPE vs.\ Standard LoRA (fil)} & BLEU & $-$2.68 & [$-$3.69, $-$1.77] & $-$1.72 & 9/9 \\*
 & chrF++ & $-$3.67 & [$-$4.55, $-$2.64] & $-$2.35 & 9/9 \\*
 & GLEU & $-$2.44 & [$-$3.13, $-$1.70] & $-$2.09 & 9/9 \\
\cmidrule(lr){1-6}
\multirow{3}{*}{Exp 3 LAPE vs.\ Standard LoRA (eng)} & BLEU & $-$3.59 & [$-$5.08, $-$2.04] & $-$1.45 & 8/9 \\*
 & chrF++ & $-$4.67 & [$-$6.12, $-$3.17] & $-$1.97 & 8/9 \\*
 & GLEU & $-$3.23 & [$-$4.72, $-$1.69] & $-$1.33 & 8/9 \\
\midrule
\multirow{3}{*}{Exp 1 LAPE vs.\ Random Mask} & BLEU & 0.29 & [$-$0.82, 1.24] & 0.17 & 6/9 \\*
 & chrF++ & 0.93 & [$-$0.23, 2.06] & 0.49 & 6/9 \\*
 & GLEU & 0.53 & [$-$0.63, 1.45] & 0.31 & 7/9 \\
\cmidrule(lr){1-6}
\multirow{3}{*}{Exp 2a LAPE vs.\ Random Mask} & BLEU & $-$0.15 & [$-$1.16, 0.61] & $-$0.10 & 3/9 \\*
 & chrF++ & 0.32 & [$-$0.74, 1.16] & 0.20 & 7/9 \\*
 & GLEU & $-$0.11 & [$-$1.52, 1.20] & $-$0.05 & 5/9 \\
\cmidrule(lr){1-6}
\multirow{3}{*}{Exp 2b LAPE vs.\ Random Mask} & BLEU & 0.13 & [$-$1.02, 1.00] & 0.08 & 7/9 \\*
 & chrF++ & 0.93 & [$-$0.41, 2.07] & 0.46 & 8/9 \\*
 & GLEU & 0.25 & [$-$0.99, 1.10] & 0.14 & 7/9 \\
\cmidrule(lr){1-6}
\multirow{3}{*}{Exp 3 LAPE vs.\ Random Mask} & BLEU & $-$0.25 & [$-$1.15, 0.56] & $-$0.18 & 4/9 \\*
 & chrF++ & $-$0.33 & [$-$1.19, 0.39] & $-$0.25 & 5/9 \\*
 & GLEU & $-$0.63 & [$-$1.60, 0.37] & $-$0.39 & 6/9 \\
\midrule
\multirow{3}{*}{Pan-AN vs.\ English (2b vs.\ 3)} & BLEU & $-$0.57 & [$-$1.27, 0.15] & $-$0.48 & 5/9 \\*
 & chrF++ & $-$1.11 & [$-$2.53, 0.43] & $-$0.46 & 6/9 \\*
 & GLEU & $-$0.61 & [$-$2.04, 0.92] & $-$0.25 & 5/9 \\
\cmidrule(lr){1-6}
\multirow{3}{*}{Pan-AN vs.\ Filipino (2b vs.\ 2a)} & BLEU & 0.31 & [$-$0.22, 1.06] & 0.29 & 5/9 \\*
 & chrF++ & 0.50 & [$-$0.40, 1.80] & 0.27 & 5/9 \\*
 & GLEU & 0.28 & [$-$0.51, 1.38] & 0.18 & 4/9 \\
\cmidrule(lr){1-6}
\multirow{3}{*}{Filipino vs.\ Yami (2a vs.\ 1)} & BLEU & $-$4.56 & [$-$5.11, $-$3.86] & $-$4.44 & 9/9 \\*
 & chrF++ & $-$5.47 & [$-$6.28, $-$4.63] & $-$4.01 & 9/9 \\*
 & GLEU & $-$5.07 & [$-$5.85, $-$4.23] & $-$3.85 & 9/9 \\
\midrule
\multirow{3}{*}{Yami vs.\ Neg.\ Ctrl (1 vs.\ 0)} & BLEU & 0.72 & [$-$0.11, 1.67] & 0.50 & 5/9 \\*
 & chrF++ & 1.28 & [0.06, 2.64] & 0.61 & 5/9 \\*
 & GLEU & 0.92 & [0.20, 1.62] & 0.79 & 6/9 \\
\cmidrule(lr){1-6}
\multirow{3}{*}{Filipino vs.\ Neg.\ Ctrl (2a vs.\ 0)} & BLEU & $-$3.83 & [$-$5.01, $-$2.56] & $-$1.92 & 9/9 \\*
 & chrF++ & $-$4.19 & [$-$6.16, $-$2.13] & $-$1.27 & 8/9 \\*
 & GLEU & $-$4.16 & [$-$5.52, $-$2.77] & $-$1.86 & 9/9 \\
\cmidrule(lr){1-6}
\multirow{3}{*}{Pan-AN vs.\ Neg.\ Ctrl (2b vs.\ 0)} & BLEU & $-$3.52 & [$-$5.02, $-$1.71] & $-$1.30 & 8/9 \\*
 & chrF++ & $-$3.69 & [$-$6.18, $-$0.70] & $-$0.82 & 8/9 \\*
 & GLEU & $-$3.87 & [$-$5.72, $-$1.81] & $-$1.22 & 8/9 \\
\cmidrule(lr){1-6}
\multirow{3}{*}{English vs.\ Neg.\ Ctrl (3 vs.\ 0)} & BLEU & $-$2.95 & [$-$3.95, $-$1.56] & $-$1.47 & 8/9 \\*
 & chrF++ & $-$2.58 & [$-$4.30, $-$0.26] & $-$0.76 & 8/9 \\*
 & GLEU & $-$3.26 & [$-$4.22, $-$2.06] & $-$1.83 & 8/9 \\
\bottomrule
\end{longtable}
\endgroup

\begin{table}[h]
\centering\small
\caption{Sign agreement across the three metrics: number of the nine
models whose model-level difference shares the sign of the overall
estimate.}
\label{tab:significance-matrix}
\begin{tabular}{lccc}
\toprule
Comparison & BLEU & chrF++ & GLEU \\
\midrule
Exp 0 LAPE vs.\ Standard LoRA (cmn) & 9/9 & 9/9 & 9/9 \\
Exp 1 LAPE vs.\ Standard LoRA (cmn) & 9/9 & 9/9 & 9/9 \\
Exp 2a LAPE vs.\ Standard LoRA (fil) & 9/9 & 9/9 & 9/9 \\
Exp 2b LAPE vs.\ Standard LoRA (fil) & 9/9 & 9/9 & 9/9 \\
Exp 3 LAPE vs.\ Standard LoRA (eng) & 8/9 & 8/9 & 8/9 \\
\midrule
Exp 1 LAPE vs.\ Random Mask & 6/9 & 6/9 & 7/9 \\
Exp 2a LAPE vs.\ Random Mask & 3/9 & 7/9 & 5/9 \\
Exp 2b LAPE vs.\ Random Mask & 7/9 & 8/9 & 7/9 \\
Exp 3 LAPE vs.\ Random Mask & 4/9 & 5/9 & 6/9 \\
\midrule
Pan-AN vs.\ English (2b vs.\ 3) & 5/9 & 6/9 & 5/9 \\
Pan-AN vs.\ Filipino (2b vs.\ 2a) & 5/9 & 5/9 & 4/9 \\
Filipino vs.\ Yami (2a vs.\ 1) & 9/9 & 9/9 & 9/9 \\
\midrule
Yami vs.\ Neg.\ Ctrl (1 vs.\ 0) & 5/9 & 5/9 & 6/9 \\
Filipino vs.\ Neg.\ Ctrl (2a vs.\ 0) & 9/9 & 8/9 & 9/9 \\
Pan-AN vs.\ Neg.\ Ctrl (2b vs.\ 0) & 8/9 & 8/9 & 8/9 \\
English vs.\ Neg.\ Ctrl (3 vs.\ 0) & 8/9 & 8/9 & 8/9 \\
\bottomrule
\end{tabular}

\end{table}

\paragraph{What the bootstrap rows mean for \S\ref{sec:rig}.} The five
LAPE-vs-Standard rows are all negative with stability intervals below
zero; on BLEU, $d_z$ ranges from $-2.64$ (Exp 1) to $-1.45$ (Exp 3)
($-3.48$ to $-1.33$ across all three metrics),
and 9 of 9 models agreeing in sign (8 of 9 for Exp 3); LAPE-masked LoRA
underperforms unconstrained Standard LoRA on every condition. The four
LAPE-vs-Random rows have $|d_z|\le 0.49$ across metrics ($|d_z|\le 0.18$
on BLEU), every stability interval includes zero, and sign agreement
ranges from 3 to 8 of 9 models. The LAPE mask shows no demonstrated advantage over a
matched random sparse mask. The four \emph{vs.\ negative control} rows
show that the Filipino, Pan-Austronesian, and English masks (Exp 2a, 2b,
3) all \emph{underperform} the unrelated-language negative control (Exp
0), $d_z$ from $-1.30$ to $-1.92$ on BLEU with stability intervals below
zero. Yami direct adaptation (Exp 1) is not clearly separated from the
negative control: on BLEU, $\Delta = +0.72$ with a stability interval
of $[-0.11, 1.67]$ and $d_z=+0.50$; the chrF++ and GLEU intervals lie just above zero, but
only 5--6 of the nine models share the sign.

\paragraph{Short entries vs.\ sentences.} A pre-specified Exp 1 analysis
splits the benchmark into short entries (1{,}700 pairs) and sentences
(1{,}866 pairs) and tests four BLEU comparisons under Holm correction,
where corrected significance is attainable (floor $0.0156$). Standard
LoRA exceeds LAPE on both short entries ($\Delta = +4.24$, interval
$[2.89, 5.60]$, $d_z = 1.92$, 9/9 models) and sentences
($\Delta = +2.61$, $[2.09, 3.26]$, $d_z = 2.71$, 9/9), both at
Holm-adjusted $p = 0.0156$. LAPE shows no advantage over Random on
either (short entries $\Delta = -0.24$, $[-2.10, 1.36]$; sentences
$\Delta = +0.38$, $[-0.58, 1.29]$; both Holm-adjusted $p = 0.945$). The
two main results therefore do not depend on the short lexical material.

\clearpage
\section{Activation similarity analysis}\label{app:yami-dendrogram}

\subsection{Methodology}

The activation profile $A_\ell \in [0,1]^N$ for language $\ell$ at a
given layer is the per-neuron firing probability over a monolingual
corpus (FineWeb-2 for the contrast languages, capped at 100M tokens;
FineWeb for English; for Yami, all Yami-side text of the training split,
about 335K tokens). Because the Yami profile comes from the training
data's Yami side, corpus composition and language identity are not
separated for Yami. Stacked across layers,
each language's profile is a $[L, N]$ tensor where $L$ is the number of
transformer layers and $N$ the MLP intermediate width.

For each language pair $(\ell, \ell')$ we compute cosine similarity
\emph{per layer} (along the neuron axis) and then take the mean across
layers,
\begin{equation}\label{eq:per-layer-cosine}
\mathrm{sim}(\ell, \ell') \;=\; \frac{1}{L}\sum_{i=1}^{L}\;
\frac{A_\ell^{(i)} \cdot A_{\ell'}^{(i)}}{\|A_\ell^{(i)}\|\;\|A_{\ell'}^{(i)}\|}.
\end{equation}
Similarity is computed layer by layer and then averaged, so that each
layer contributes equally.

\subsection{Per-model Yami nearest neighbors}

\begin{table}[htbp]
\centering
\caption{Per-model Yami nearest neighbors. Cosine similarity computed
per-layer between activation-probability profiles, then averaged across layers.
The last column
flags whether every Austronesian language (Cebuano, Filipino, Indonesian,
Maori, Plateau Malagasy) is closer to Yami than every non-Austronesian
language tested. The encoder--decoder NLLB-200 is the partial exception:
Filipino, Cebuano, and Plateau Malagasy remain Yami's nearest neighbors, but
English and Korean rank above Maori and Indonesian.}
\label{tab:yami_neighbors}
\small
\setlength{\tabcolsep}{4pt}
\begin{tabular}{l p{0.42\linewidth} l c}
\toprule
Model & Top-5 nearest neighbors (cosine) & Closest non-AN & All-AN tighter? \\
\midrule
    Qwen3-4B & Plateau Malagasy (0.890), Cebuano (0.890), Filipino (0.888), Maori (0.856), Indonesian (0.821) & Arabic (0.797) & yes \\
    Qwen3-14B & Filipino (0.867), Plateau Malagasy (0.866), Cebuano (0.864), Maori (0.840), Indonesian (0.797) & Arabic (0.778) & yes \\
    Qwen3-32B & Filipino (0.880), Cebuano (0.876), Plateau Malagasy (0.864), Maori (0.840), Indonesian (0.808) & Arabic (0.786) & yes \\
    Gemma-3-4B & Cebuano (0.789), Filipino (0.785), Plateau Malagasy (0.764), Maori (0.761), Indonesian (0.726) & Mandarin (0.707) & yes \\
    Gemma-3-12B & Cebuano (0.807), Filipino (0.801), Plateau Malagasy (0.774), Maori (0.773), Indonesian (0.722) & Mandarin (0.692) & yes \\
    Gemma-3-27B & Cebuano (0.777), Filipino (0.774), Plateau Malagasy (0.764), Maori (0.748), Indonesian (0.697) & Mandarin (0.681) & yes \\
    SeaLLMs-v3-7B & Plateau Malagasy (0.889), Cebuano (0.862), Maori (0.836), Filipino (0.834), Indonesian (0.766) & Arabic (0.723) & yes \\
    SEA-LION-v4-27B & Cebuano (0.795), Filipino (0.795), Plateau Malagasy (0.771), Maori (0.765), Indonesian (0.714) & Mandarin (0.695) & yes \\
    NLLB-200-3.3B & Filipino (0.896), Cebuano (0.878), Plateau Malagasy (0.779), English (0.738), Korean (0.729) & English (0.738) & no \\
\bottomrule
\end{tabular}
\end{table}

\begin{figure}[h]
  \centering
  \includegraphics[width=0.85\linewidth]{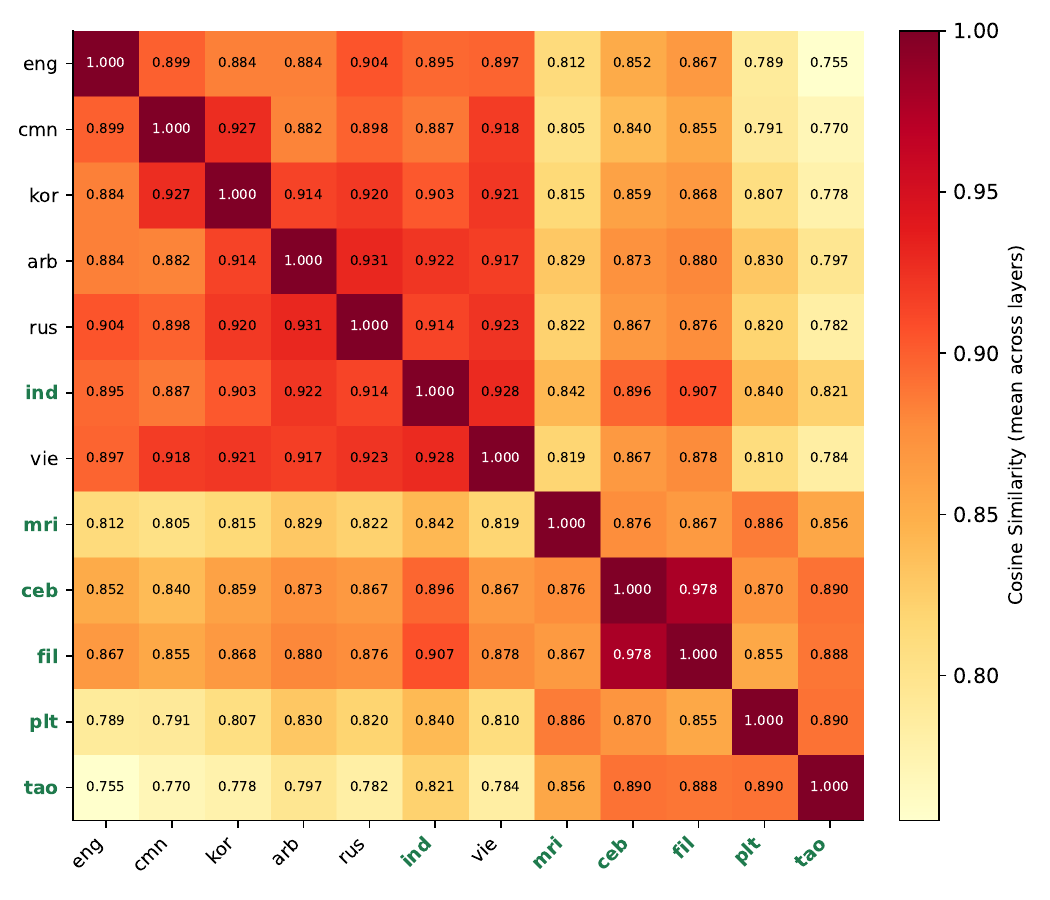}
  \caption{Pairwise activation cosine similarity heatmap (Qwen3-4B;
  per-layer cosine, mean across layers). Yami's highest similarities are
  to the Austronesian languages (Plateau Malagasy, Cebuano, Filipino,
  Maori, Indonesian); Mandarin, English, and the contrast set
  (Russian/Vietnamese/Korean/Arabic) are distal.}
  \label{fig:lang-sim}
\end{figure}

\begin{figure}[h]
  \centering
  \includegraphics[width=\linewidth]{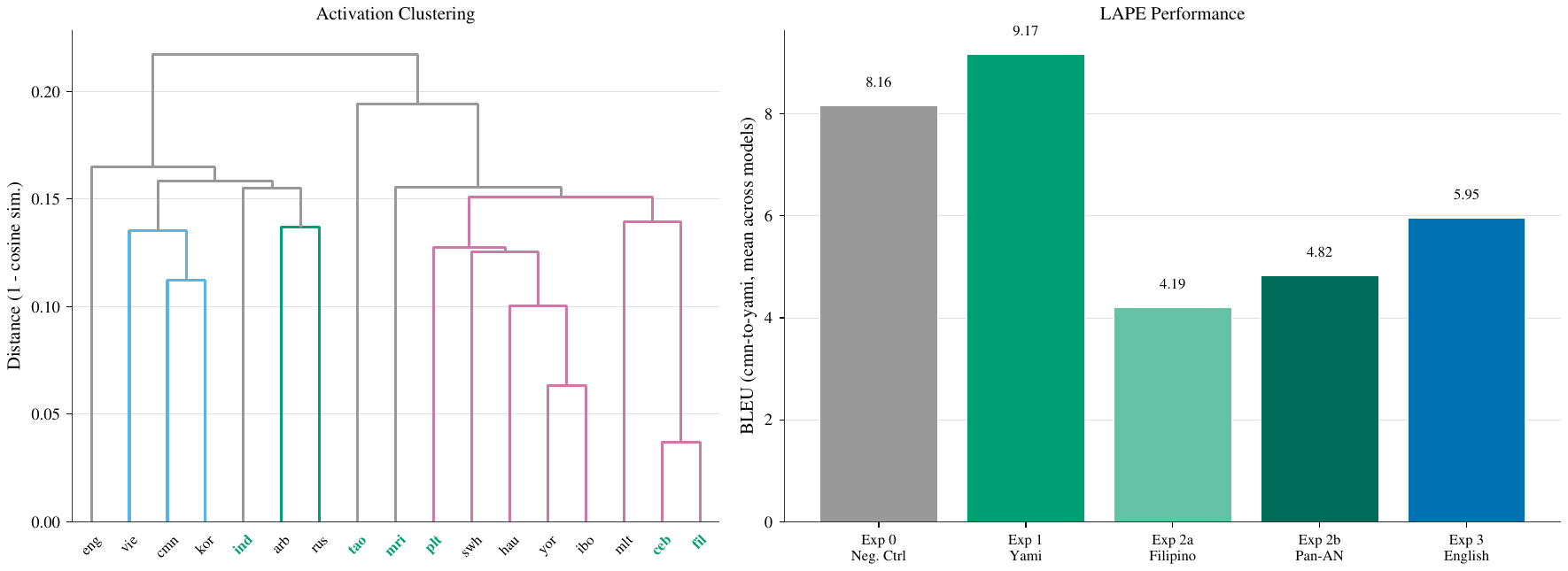}
  \caption{\emph{Left}: hierarchical clustering (average linkage) of the
  sixteen non-Yami evaluation languages and Yami (\texttt{tao}) by
  activation distance ($1-$cosine similarity of per-layer activation
  profiles, averaged across layers), averaged across all nine models.
  The low-resource languages, including most of the Austronesian set
  (green labels) and the low-resource non-Austronesian controls, group
  with Yami, while the higher-resource languages (among them
  Austronesian Indonesian, \texttt{ind}) form a separate cluster.
  \emph{Right}: mean Chinese$\to$Yami BLEU for each LAPE condition,
  averaged across models. The clustering is not matched by a
  corresponding translation gain from LAPE targeting: the Filipino and
  Pan-AN conditions fall well below the negative control, and direct
  Yami targeting (Exp 1) exceeds it only slightly; the gap holds
  whether the clustering reflects Austronesian genealogy, corpus-resource
  scale, or a mixture of the two \citep{lian2026dissertation}.}
  \label{fig:yami-dendrogram}
\end{figure}

Figure~\ref{fig:lang-sim} shows the similarity matrix for Qwen3-4B,
and Figure~\ref{fig:yami-dendrogram} the nine-model clustering including
the low-resource controls. In every
one of the nine models tested, Yami's three nearest neighbors
are Austronesian, drawn from \{Filipino, Cebuano, Plateau Malagasy,
Maori, Indonesian\}. In eight of nine, the entire Austronesian set is
closer to Yami than every non-Austronesian language tested. The
encoder--decoder NLLB-200 is the partial exception: Filipino, Cebuano,
and Plateau Malagasy remain Yami's nearest neighbors, but English and
Korean rank above Maori and Indonesian. In the NLLB checkpoint, the
\texttt{tao\_Latn} language token was initialized by cloning the Tagalog
token and kept fixed, which bears on its proximity to Filipino. Mandarin
is not uniformly distal: it is
the closest non-Austronesian language to Yami in the three Gemma~3 models
and SEA-LION-v4, whereas Arabic is closest in the Qwen3 models and
SeaLLMs-v3. These neighbor rankings cover only the eleven comparison
languages: when five low-resource non-Austronesian controls are added,
three of them (Yoruba, Igbo, Hausa) fall within the Austronesian
distance range, and distance from Yami tracks log FineWeb-2 corpus size,
a public proxy for pretraining exposure (Pearson $r = 0.879$ across
sixteen languages), more than genealogy
\citep{lian2026dissertation}.

\clearpage
\section{Mask overlap and the Mandarin-collateral-damage test}\label{app:overlap}

\subsection{Inter-mask Jaccard}

\begin{figure}[h]
  \centering
  \includegraphics[width=0.95\linewidth]{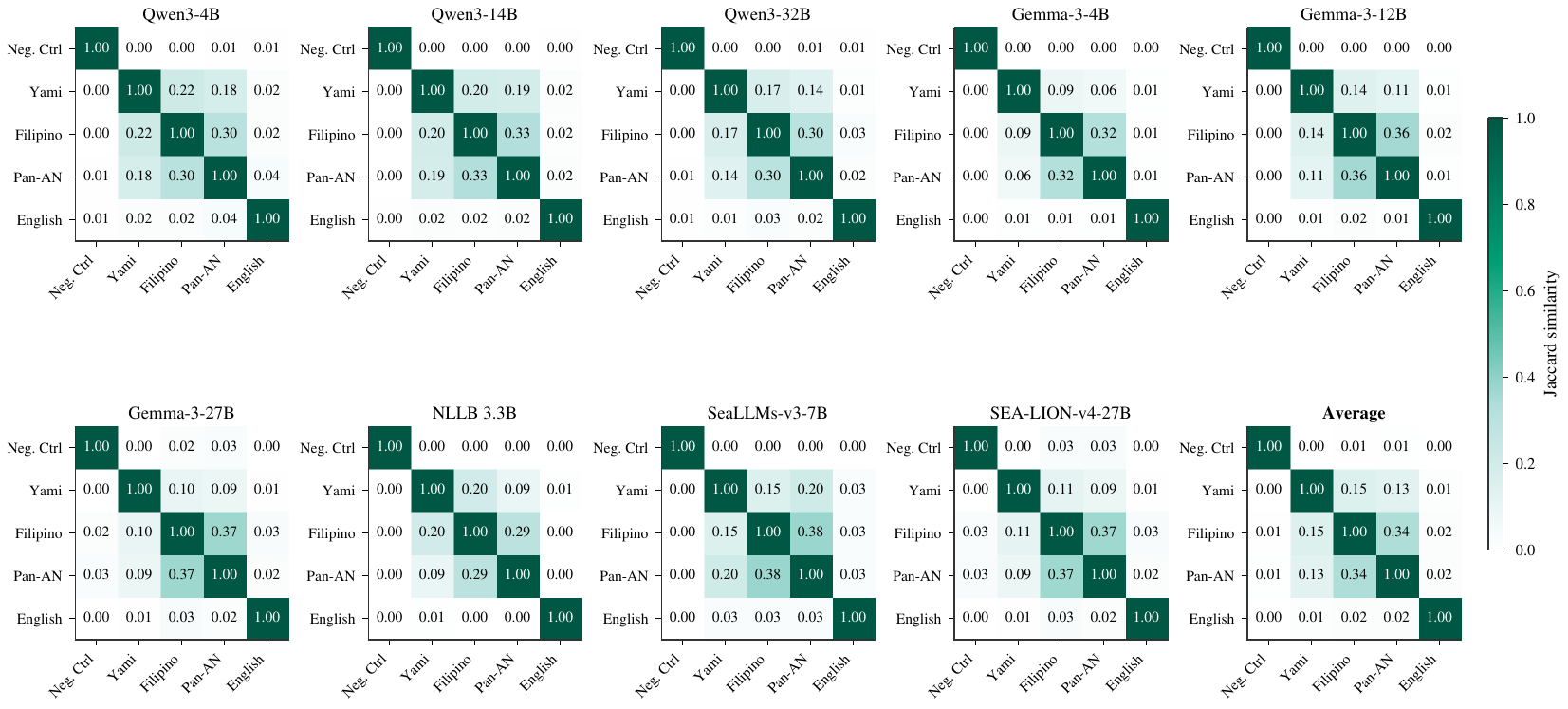}
  \caption{Per-(model, mask-pair) Jaccard heatmap. The (Yami, Neg.\
  Ctrl) cell is uniformly near zero across all nine models.}
  \label{fig:jaccard-heatmap}
\end{figure}

Pairwise Jaccard between the experimental masks themselves
(Table~\ref{tab:mask_overlap}, Figure~\ref{fig:jaccard-heatmap})
verifies that they are largely disjoint. The negative-control mask
(Exp 0) shares essentially no neurons with Yami (mean Jaccard $0.000$);
the strongest within-experiment overlap is between Filipino and
Pan-Austronesian ($0.336$), as expected since Filipino is one of the
five Pan-AN target languages. The negative control's competitive
performance therefore cannot be attributed to mask overlap with Yami.

\begin{table}[htbp]
\centering
\caption{Pairwise Jaccard similarity between experimental masks, averaged across all 9 models.
  Values close to 0 indicate distinct neuron populations; values close to 1 indicate
  near-identical masks.}
\label{tab:mask_overlap}
\small
\begin{tabular}{l ccccc}
\toprule
 & Neg. Ctrl & Yami & Filipino & Pan-AN & English \\
\midrule
    Neg. Ctrl & 1.000 & 0.000 & 0.007 & 0.009 & 0.003 \\
    Yami & 0.000 & 1.000 & 0.152 & 0.128 & 0.013 \\
    Filipino & 0.007 & 0.152 & 1.000 & 0.336 & 0.020 \\
    Pan-AN & 0.009 & 0.128 & 0.336 & 1.000 & 0.018 \\
    English & 0.003 & 0.013 & 0.020 & 0.018 & 1.000 \\
\bottomrule
\end{tabular}
\end{table}

\subsection{Mandarin-collateral-damage test}\label{app:cmn-overlap}

A natural worry about the genealogical inversion documented in
\S\ref{sec:rig} is that the Filipino and Pan-Austronesian LAPE masks
might \emph{accidentally} overlap with neurons that participate in
Mandarin processing, which the Yami$\leftrightarrow$Chinese benchmark
requires even though the cross-pair training data contain no Mandarin.
In that case, the Filipino-LAPE underperformance would reflect
collateral damage to Mandarin processing rather than failure of
genealogical localization.

The hypothesis is testable directly: LAPE also identifies a
Mandarin-specific neuron mask in each model. Computing per-mask Jaccard
overlap with this Mandarin mask, averaged across the nine models
(Table~\ref{tab:cmn_overlap}), yields the opposite of what the
collateral-damage hypothesis predicts.

\begin{table}[htbp]
\centering
\caption{Overlap between selected masks and the model's LAPE-identified
Mandarin mask, averaged across the nine models. This tests whether the
proxy-language masks may be harming the common Yami$\leftrightarrow$Chinese
benchmark by masking Mandarin-associated neurons. The proxy-language masks
overlap it far less than the negative-control union does. Korean,
Vietnamese, and Russian are separate single-language masks shown for reference.}
\label{tab:cmn_overlap}
\small
\begin{tabular}{l r r}
\toprule
Mask & Mean Jaccard & Mean shared neurons \\
\midrule
    Korean & 0.1088 & 138.2 \\
    Vietnamese & 0.0783 & 102.0 \\
    Russian & 0.0233 & 26.9 \\
    Neg.\ Ctrl (union) & 0.0100 & 51.0 \\
    English & 0.0014 & 7.3 \\
    Filipino & 0.0003 & 3.8 \\
    Yami & 0.0003 & 3.3 \\
    Pan-AN & 0.0001 & 1.1 \\
\bottomrule
\end{tabular}
\end{table}

The Filipino and Pan-Austronesian masks overlap the Mandarin mask
\emph{less} than the unrelated-language negative-control union does
(mean Jaccard $0.0003$ and $0.0001$ respectively, vs.\ $0.0100$ for the
Russian/Vietnamese/Korean union). The Korean, Vietnamese, and Russian
rows are separate single-language LAPE masks, shown for reference. If
Mandarin collateral damage explained the inversion, the
unrelated-language condition should perform worst; instead it
outperforms all three proxy-language conditions.

\paragraph{What this test does and does not rule out.} The Jaccard test
weighs against one specific operationalization of ``Mandarin interference'':
overlap between the experimental mask and the LAPE-identified
Mandarin-specific neuron set. A skeptic could still argue that
Mandarin processing is not exhaustively captured by the Mandarin mask
(high-resource languages may distribute their processing across
neurons that LAPE does not flag as \emph{specific} to Mandarin). The
Jaccard test does not rule out such distributed accounts; it only weighs
against the simplest, most obvious version of the worry. The ordering
also remains confounded with the training pair and its provenance: the
proxy-language conditions train on pivot-translated Yami--Filipino or
Yami--English data, the negative control on the human-translated
Yami--Chinese benchmark pair, and no ablation separates these factors.
Finally, the negative-control set is unrelated to Yami but not free of
Sinitic influence: Korean has a Sino-Korean vocabulary stratum
\citep{lee2011history} and Vietnamese an Early Sino-Vietnamese layer
\citep{alves2016early}, and contact history with Chinese was not
controlled. The ordering is therefore reported descriptively.

\clearpage
\section{Per-model failure-mode atlas}\label{app:atlas}

The aggregate metric numbers obscure systematic per-model qualitative
behavior. Three tables document the most diagnostic patterns.

\subsection{Default-vocabulary fingerprints under cross-pair LAPE}

When Filipino-LAPE is trained on Yami$\leftrightarrow$Filipino and
evaluated on Yami$\to$Chinese (cross-pair, the direction where the
0.000-BLEU collapse is most pronounced), each of the eight decoder-only
models has a most-frequent Chinese phrase that recurs across many
different Yami sources, usually an everyday or maritime word such as
``small fish'', ``boat'', or ``conch shell''
(Table~\ref{tab:model-fingerprints-collapse}).

\begin{table}[htbp]
\centering
\caption{The Yami$\to$Chinese collapse defaults to a small Yami-domain Chinese vocabulary (English glosses shown), and the default token differs by model. Filipino-LAPE condition (Exp 2a), all 3{,}566 test items per model. \textbf{NLLB} differs qualitatively: it emits whitespace rather than a default token.}
\label{tab:model-fingerprints-collapse}
\small
\begin{tabular}{l l r r r}
\toprule
Model & Top-1 default prediction & Count & Unique preds & Empty preds \\
\midrule
Qwen3-4B        & ``small fish''       & 125 & 2652 & 0   \\ %
Qwen3-14B       & ``hair''             & 137 & 2824 & 0   \\ %
Qwen3-32B       & ``small boat''       & 270 & 2811 & 0   \\ %
Gemma-3-4B      & ``very difficult''   &  57 & 2751 & 0   \\ %
Gemma-3-12B     & ``hunting''          &  34 & 2861 & 0   \\ %
Gemma-3-27B    & ``catch by hand''    &  45 & 2896 & 0   \\ %
SeaLLMs-v3-7B   & ``small boat''       &  86 & 2678 & 0   \\ %
SEA-LION-v4-27B & ``conch shell''      &  64 & 2958 & 0   \\ %
NLLB-200-3.3B   & \textit{(whitespace)} & 406 & 2677 & 406 \\
\bottomrule
\end{tabular}
\end{table}

\subsection{Pan-Austronesian per-model breakdown}

Pan-AN (Exp 2b) cross-pair degeneration is bimodal across models
(Table~\ref{tab:model-fingerprints-panan}): some models produce long
streams of Yami particles with no lexical content; others leak Chinese
characters into the Yami output. SEA-LION-v4 leaks the most CJK
despite SEA-language pretraining; Gemma-3-27B is the most stable. These
over-generation patterns could come from the attention adapters, the
sparse MLP constraint, the training-pair mismatch, or their
interactions; the outputs alone cannot separate these sources.

\begin{table}[htbp]
\centering
\caption{Pan-AN (Exp 2b) Chinese$\to$Yami degeneration is bimodal: some models produce long \textit{Yami particle streams} (no lexical content); others leak Chinese characters into the Yami output. SEA-LION-v4 leaks the most CJK despite SEA-language pretraining; Gemma-3-27B is the most stable.}
\label{tab:model-fingerprints-panan}
\small
\begin{tabular}{l r r r l r}
\toprule
Model & Long ($>$50 wc) & Max wc & CJK leak & Top token in long preds & ($\times$) \\
\midrule
Qwen3-4B & 184 & 256 & 23 & \texttt{ya} & 2810 \\
Qwen3-14B & 121 & 191 & 1 & \texttt{a} & 2003 \\
Qwen3-32B & 29 & 253 & 2 & \texttt{mo} & 403 \\
Gemma-3-4B & 70 & 254 & 73 & \texttt{a} & 2080 \\
Gemma-3-12B & 55 & 256 & 23 & \texttt{am,} & 913 \\
Gemma-3-27B & 7 & 170 & 20 & \texttt{a} & 197 \\
SeaLLMs-v3-7B & 135 & 256 & 3 & \texttt{a} & 2431 \\
SEA-LION-v4-27B & 41 & 249 & 86 & \texttt{a} & 1351 \\
NLLB-200-3.3B & 14 & 255 & 0 & \texttt{among} & 458 \\
\bottomrule
\end{tabular}
\end{table}

\subsection{NLLB direction asymmetry}

NLLB-200 produces 2{,}928 of the 2{,}929 empty predictions in the
entire dataset (the remaining one is a single Qwen3-14B run). The
NLLB empties concentrate sharply on Yami$\to$Chinese under cross-pair
sparse-mask (LAPE or Random) training (Table~\ref{tab:nllb-empty-by-condition}); on
Chinese$\to$Yami the same conditions produce 0--8 empties. The empty
mode arises under the combination of cross-pair training, sparse
masking (LAPE \emph{or} Random), and short source inputs; Standard LoRA
on the same training data produces $\le 1\%$ empties.

\begin{table}[htbp]
\centering
\caption{NLLB Yami$\to$Chinese empty-prediction rate by condition (3{,}566 items per condition). Empty outputs are concentrated in the cross-pair conditions under sparse masking (LAPE \emph{or} Random); the matched-pair and Standard-LoRA conditions stay at or below 1\%.}
\label{tab:nllb-empty-by-condition}
\small
\begin{tabular}{l r r}
\toprule
Condition & Empty preds / total & Rate \\
\midrule
Exp 1 (Yami), Random Mask & 0 / 3566 & 0.0\% \\
Standard LoRA, Yami--Chinese & 0 / 3566 & 0.0\% \\
Exp 0 (Neg.\ Ctrl), LAPE & 0 / 3566 & 0.0\% \\
Exp 1 (Yami), LAPE & 0 / 3566 & 0.0\% \\
Standard LoRA, Yami--English & 4 / 3566 & 0.1\% \\
Base model (no fine-tuning) & 9 / 3566 & 0.3\% \\
Standard LoRA, Yami--Filipino & 35 / 3566 & 1.0\% \\
Exp 2b (Pan-AN), LAPE & 118 / 3566 & 3.3\% \\
Exp 2a (Filipino), LAPE & 406 / 3566 & 11.4\% \\
Exp 2b (Pan-AN), Random Mask & 465 / 3566 & 13.0\% \\
Exp 2a (Filipino), Random Mask & 497 / 3566 & 13.9\% \\
Exp 3 (English), LAPE & 661 / 3566 & 18.5\% \\
Exp 3 (English), Random Mask & 705 / 3566 & 19.8\% \\
\bottomrule
\end{tabular}
\end{table}

\clearpage
\section{Code and data availability}\label{app:repro}

We do not release code or data with this version of the paper. The
Yami--Mandarin parallel corpus used in \S\ref{sec:rig} was compiled
for the second author's dissertation and is sourced from third-party
data providers (D.~Victoria Rau and Ann Chang); the corpus itself is not the authors' to release, and raw
predictions and reference translations are withheld for the same
reason. The dissertation
\citep{lian2026dissertation} provides full methodological detail and
the complete experimental record.

\end{document}